\documentclass[nohyperref]{article}
\usepackage{microtype}
\usepackage{graphicx}
\usepackage{float}
\usepackage{subfigure}
\usepackage{scalerel}
\usepackage{booktabs} 

\usepackage{hyperref}

\usepackage[accepted]{icml2023}

\usepackage{amsmath}
\usepackage{amssymb}
\usepackage{mathtools}
\usepackage{amsthm}
\usepackage{notation}
\usepackage{dsfont}

\usepackage[disable,textsize=tiny]{todonotes}

\icmltitlerunning{Training Normalizing Flows with the Precision-Recall Divergence}

\begin{document}

\twocolumn[
\icmltitle{Training Normalizing Flows with the Precision-Recall Divergence}



\icmlsetsymbol{equal}{*}

\begin{icmlauthorlist}
\icmlauthor{Alexandre Verine}{lam}
\icmlauthor{Benjamin Negrevergne}{lam}
\icmlauthor{Muni Sreenivas Pydi}{lam}
\icmlauthor{Yann Chevaleyre}{lam}

\end{icmlauthorlist}

\icmlaffiliation{lam}{LAMSADE, CNRS, Universit\'e Paris-Dauphine-PSL, Paris, France}
\icmlcorrespondingauthor{Alexandre Verine}{alexandre.verine@dauphine.psl.eu}

\icmlkeywords{Machine Learning, Normalizing Flow, ICML, Precision Recall}
\vskip 0.3in
]



\printAffiliationsAndNotice{}  

\begin{abstract}

    Generative models can have distinct mode of failures like mode dropping and low quality samples, which cannot be captured by a single scalar metric. To address this, recent works propose evaluating generative models using precision and recall, where precision measures quality of samples and recall measures the coverage of the target distribution. 
    Although a variety of discrepancy measures between the target and estimated distribution are used to train generative models, it is unclear what precision-recall trade-offs are achieved by various choices of the discrepancy measures. 
    In this paper, we show that achieving a specified precision-recall trade-off corresponds to minimising $f$-divergences from a family we call the {\em PR-divergences }. Conversely, any $f$-divergence can be written as a linear combination of PR-divergences and therefore correspond to minimising a weighted precision-recall trade-off. Further, we propose a novel generative model that is able to train a normalizing flow to minimise any $f$-divergence, and in particular, achieve a given precision-recall trade-off.


\end{abstract}

\section{Introduction}

Generative models have emerged as a powerful tool in machine learning. In recent years, a plethora of generative models have been proposed, such as Generative Adversarial Networks (GANs) and Normalizing Flows, which have shown exceptional performance in learning complex, high-dimensional distributions on image and text data.
Despite the successes, evaluation of generative models remains a tricky problem. While approaches based on GANs often produce high fidelity images, they suffer from mode dropping (inability to produce data from all the modes of the true distribution) and poor likelihoods. Models based on normalizing flows are explicitly trained for achieving high likelihoods, but often produce low quality samples. To address these complexities, recent works have proposed a two-fold evaluation  which considers both precision and recall~\citep{sajjadi_assessing_2018, simon_revisiting_2019}. 

Given a target distribution $P$ and a set of parameterised distributions $\{P_\theta | \theta\in \Theta\}$, a generative model tries to find the best fit $\wh P = P_{\theta^*}$ that minimises  a divergence or distance metric between $P$ and $\wh P$. Typically, normalizing flows are trained by minimising the Kullback–Leibler divergence (denoted by $\KL$) between $P$ and $\wh P$. The generators in GANs can be trained with a variety of divergences / distances, for example, any $f$-divergence (denoted by $D_f$) \citep{nowozin_f-gan_2016} or the Wasserstein distance \citep{arjovsky_wasserstein_2017}. It is known that optimising the (forward) $\KL$ tends to favour {\em mass-covering} models \cite{minka_divergence_2005} which contrast with the {\em mode seeking} behaviour that we observe in  most other generative models. As illustrated in Figure~\ref{fig:1DPR}, this results in models with good recall (more diverse examples), at the cost of a lower precision (more outliers). However, it is unclear what trade-offs are made implicitly by optimising for a general divergence. This motivates the following question. 
\begin{question}\label{question: D lambda PR}
What precision-recall trade-off does minimising $D_f(P \Vert \wh P)$ correspond to?
\end{question}
Intuitively, precision of a generative model measures the quality of the samples produced and recall measures how well the target distribution is covered. Depending on the application, it may be desirable to train a generative model for high precision (for example, image synthesis and data augmentation) or high recall (for example, density estimation and denoising). However existing generative modelling approaches do not afford such flexibility. Although several works exist on efficient evaluation of the precision-recall curves of existing models, there seems to be little work on approaches for training generative models that can target a specific precision-recall trade-off. This leads to the following question.
\begin{question}\label{question: how to train model}
Is it possible to train a generative model that achieves a specified trade-off between precision and recall?
\end{question}

\begin{figure*}[ht]\label{fig: first page}
    \centering
    \subfigure[ $\lambda = 0.1$]{\includegraphics[width=0.22\textwidth]{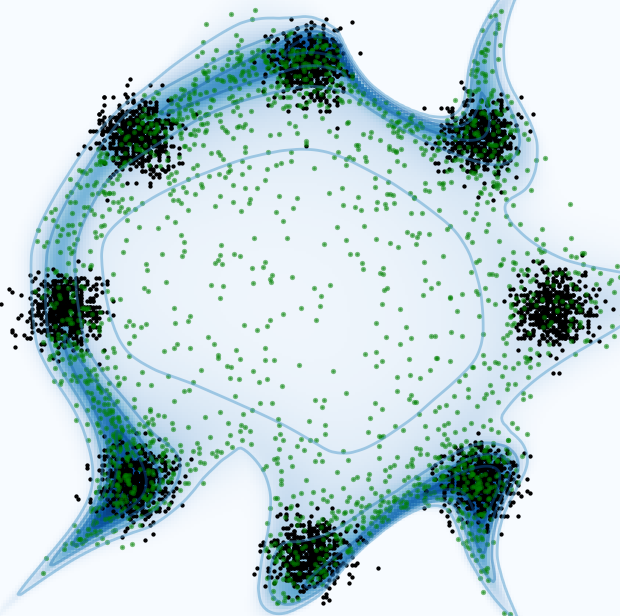} \label{fig:l01}}
    \subfigure[ $\lambda = 1$]{\includegraphics[width=0.22\textwidth]{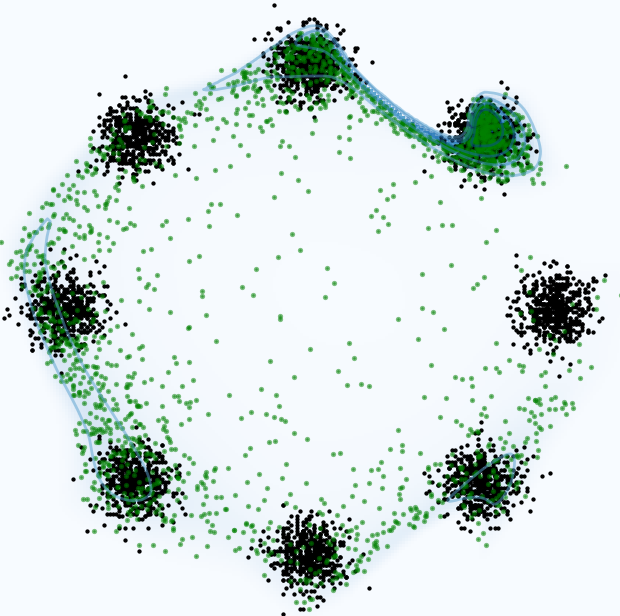}\label{fig:l1}}
    \subfigure[ $\lambda = 10$]{\includegraphics[width=0.22\textwidth]{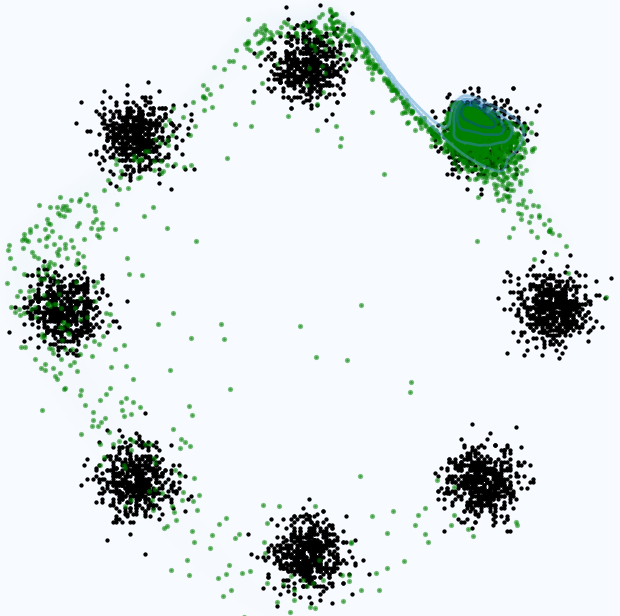}\label{fig:l10}}
    \subfigure[PR Curves]{\includegraphics[width = 0.24\textwidth]{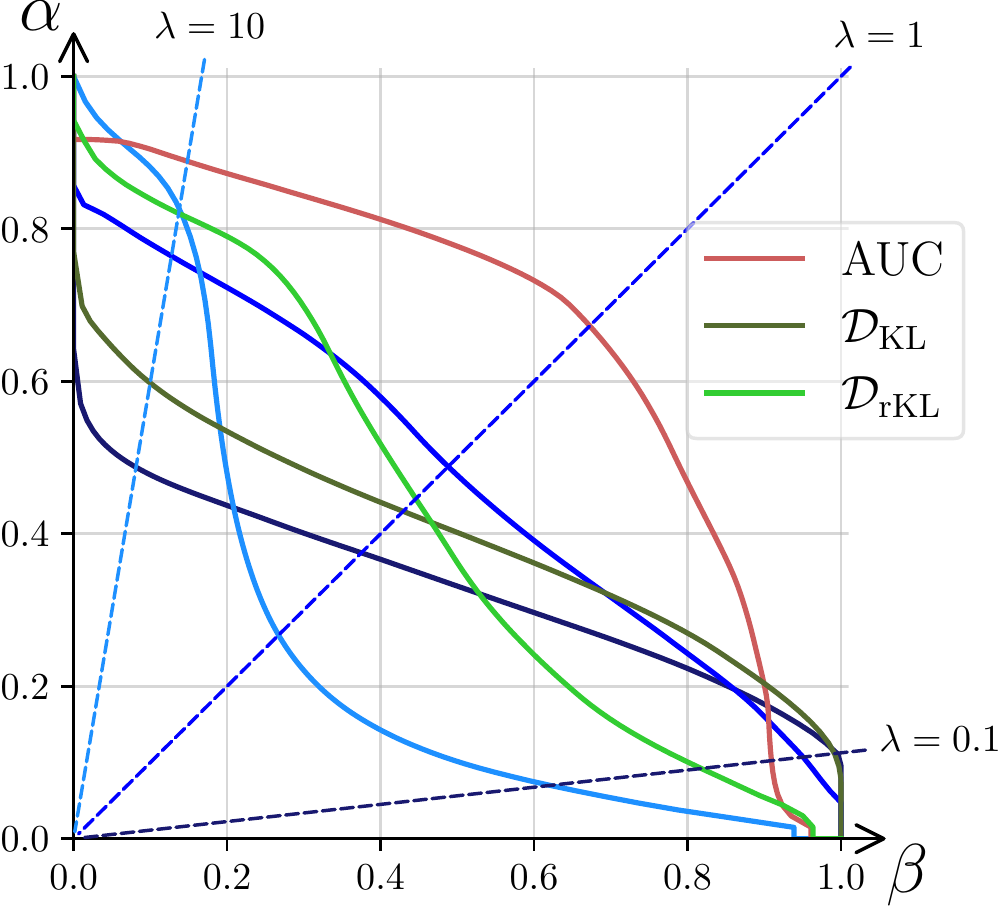} \label{fig:PRGaus}}
    \caption{Normalizing Flows (RealNVP) trained on two dimensional data-points sampled from 8 Gaussians. Fig.~\ref{fig:l01} to Fig.~\ref{fig:l10}: three models  trained to minimise the Precision-Recall Divergence with a different precision/recall trade-off. Samples drawn from the true distribution $P$ are represented in black, samples from trained models $\wh P$ are represented in green, and the log-likelihood of $\whP$ is  represented as the blue levels (darker means higher density). (a) $\lambda = 0.1$, favours  recall over  precision. (b) $\lambda = 1$, balanced precision vs. recall trade-off. (c) $\lambda = 10$, favours precision over recall.         
    Fig.~\ref{fig:PRGaus}: Precision-recall curves of models trained with different divergences. In blue, the three models from Figure~\ref{fig: first page} (a) to (c). In dark and light green respectively,  models  trained using $\KL$ divergence, and $\rKL$.  In red, the model has been trained on the AUC. (See also Fig.~\ref{fig:kl}, Fig.~\ref{fig:rkl} and  Fig.~\ref{fig:auc}.)
    }
\end{figure*}

In this paper, we address Questions~\ref{question: D lambda PR} and \ref{question: how to train model} by making the following contributions.
\begin{itemize}
    \item We show that achieving a specified precision-recall trade-off corresponds to minimising a particular f-divergence between $P$ and $\wh P$. Specifically, in Theorem~\ref{thm:DPR} we give a family of $f$-divergences (denoted by $\DPR$, $\lambda\in [0, \infty]$) that are associated with various points along the precision-recall curve of the generative model.
    \item We show that any arbitrary $f$-divergence can be written as a linear combination of $f$-divergences from the $\DPR$ family. This result makes explicit, the implicit precision-recall trade-offs made by generative models that minimise an arbitrary $f$-divergence.
	\item We propose a novel flow-based generative model that can achieve a user specified trade-off between precision and recall by minimising $\DPR$ for any given $\lambda$. 
\end{itemize}
Figure~\ref{fig: first page} shows how our model performs under various settings of $\lambda$. With a high $\lambda$, we can train the model to favour precision over recall and vice-versa.

Because the discriminator trained with $\DPR$ is mostly flat away from the origin, it is not feasible to directly train an $f$-GAN to minimise $\DPR$ lest we suffer from vanishing gradients.
We solve this problem by adversarially training a normalizing flow in conjunction with a discriminator. Our model have the following novelties over existing works.
\begin{itemize}
    \item The normalizing flow of our model minimises $\DPR$ instead of the usual $\KL$. In fact, one can train a normalizing flow to minimise any $f$-divergence using our scheme.
    \item To train the discriminator of our model, we use a divergence $\D_f$ that is different from $\DPR$ in order to have stable training without vanishing gradients. This separates our training from the min-max training of  GANs and flow-GANs, as we use different loss functions for generator and discriminator. The conditions for compatibility between the two divergences used for training are stated in Theorem~\ref{thm:boundedestimation}.
\end{itemize}


\paragraph{Notation:} We use $\setP(\setX)$ to denote the set of probability measures on the space $\setX$. We use uppercase letters to denote probability measures and the corresponding lower case letters to denote their density functions. Throughout the paper, we use $\setX\subseteq \mathbb R^d$ for the input space, $P\in \setP(\setX)$ for the target distribution we want to model and $\wh P\in \setP(\setX)$ for the distribution estimated by a generative model. Further, we assume that $P$ and $\wh P$ share the same support in $\setX$.

\section{Related works}
The task of improving Normalizing Flows by tackling the trade-off between the natural effect of mass covering of $\KL$ and a desired mode seeking result has been done in some previous work. Reject can be used to artificially improve the precision of a model \cite{stimper_resampling_2022, issenhuth_latent_2022, tanielian_learning_2020}. But by only focusing on the loss, an approach of \citet{midgley_flow_2022} trains a Normalizing Flow to minimise an $\alpha$-Divergence for $\alpha=2$,  estimated with annealed importance sampling. While the work show the benefits from training with another \fdiv, it fixes the trade-off by fixing the $\alpha$. Moreover, their work requires to know the target distribution density $P$ as they aim to train the Flow to be sampler. 

To generalise the approach of training a flow on any \fdiv divergence in an unsupervised setting, our method relies on the work of \citet{nowozin_f-gan_2016} for GANs. This framework has been evaluated on Normalizing Flows by \citet{grover_flow-gan_2018}. It has been shown that Flows can be trained with a discriminator and that adversarially trained model gain from a log-likelihood terms, however the authors restrain their evaluation to Wasserstein distance only and do not explore how the \fdiv affects the Flow.

Moreover, we make the link between \fdivs and Precision and Recall. Several works have been trying to evaluate the these values with clustering methods reduced dimension space \cite{sajjadi_assessing_2018, tanielian_learning_2020, kynkaanniemi_improved_2019}. We use likelihood ratio estimation to compute the metrics like \cite{simon_revisiting_2019}, but we offer guaranties on the the quality of the estimation. Finally, while we formulate the precision as an \fdiv, \citet{djolonga_precision-recall_2020} introduce a different evaluation framework based on R\'enyi divergences. All theses previous works only focus on the evaluation of the model but we build method to train models based on this metric. 

\section{Background}

\subsection{Preliminaries on Normalising Flows}

Let $\mathcal Z \subseteq \mathbb R^d$ be a latent space. Let $Q\in \setP(\setZ)$ denote the standard normal distribution, $\mathcal N (0, \mathcal I_d)$.
Given a target distribution $P\in \setP(\setX)$, one seeks a {\em Normalizing Flow} \citep{rezende_variational_2016} i.e., a bijection $F: \setX \to \setZ$ such that $P(A) = Q(F^{-1}(A))$ for any measurable $A\subseteq \setX$. One can then use $F$ to generate samples from $P$ by applying $F^{-1}$ to samples from $Q$, or compute the density $p(x)$ for $x\in \setX$ using the change of variable formula, $p (\vx) = q(F(\vx)) \vert \det \Jac_F(\vx) \vert$, where $\det \Jac_F(\vx)$ is determinant of Jacobian matrix of $F$ at $\vx$.


In practice, $F$ is not known, and the goal is to find a  mapping $F'$ that yields the best approximation 
$\wh P_{F'}$ of $P$. To do so, we consider a class of invertible functions $\{F_\theta: \theta\in \Theta\}$ typically represented using invertible neural network (INN) architectures such as GLOW \citep{kingma_glow_2018}, RealNVP \cite{rezende_variational_2016} or  ResFlow \cite{behrmann_invertible_2019, chen_residual_2020}. We can then train the INN to minimise the KL-divergence 
between $P$ and $\wh P_{F_\theta}$, or equivalently, by maximising the log-likelihood $\E_{\vx\sim P}\left[ \log \hat p_{F_\theta}(\vx)\right]$ estimating using a dataset $\DD$ of samples from $P$.
\begin{align}
	\begin{split}
		\label{eq:kl}
	\min_{\theta}\ &\KL(P\Vert \wh P_{F_\theta})  = \min_{\theta}\int_\Xset p(\vx)\log \frac{p(\vx)}{\hat p_{F_\theta}(\vx)}\d\vx \\
	&=  H(P) - \max_{\theta} \E_{\vx\sim P}\left[ \log \hat p_{F_\theta}(\vx)\right],
	\end{split}
\end{align}
where $H(P)$ is the entropy of $P$. In the rest of this paper, we omit the dependence on $F_{\theta}$ in $\wh P_{F_\theta}$ when it is clear from context.

In order to perform well in practice, the INN must satisfy several properties \citep{kobyzev_normalizing_2020}: the forward pass and the inverse pass must be efficient to compute ; computing the determinant of the Jacobian matrix must be  tractable;  finally, $\{F_\theta: \theta\in \Theta\}$ should be expressive enough to model $P$.
While a number of architectures have been introduced to solve the first two properties, expressivity remains a challenge because of the invertibility constraint. As a consequence $\wh P \neq P$ in most practical scenarios, and thus the divergence that is used to measure the distance between $P$ and $\wh P$ has a critical impact on the resulting model.



\begin{table*}[t]
\caption{List of usual \fdivs. The generator function $f$ is given with its Fenchel Conjugate. The optimal discriminator $\Topt$ is given in order to compute $p(\vx)/\whp(\vx)$. Then $f''(1/\lambda)/\lambda^3$ is given to compute the \fdiv as combination of Precision-Recall Divergence.}
\label{tab:fdiv}
\begin{center}
	\begin{small}
\begin{sc}

\begin{tabular*}{\textwidth}{l @{\extracolsep{\fill}} ccccc}
\toprule
Divergence & Notation & $f(u)$ & $f\s(t)$ &  $\Topt(\vx)$ & $f''(1/\lambda)/\lambda^3$ \\
\midrule \addlinespace[0.5em]
KL    & $\KL(P\Vert \whP) $ & $u\log u$& $\exp(t-1)$ & $1 + \log p({\vx})/\whp(\vx)$ & $1/\lambda^2$ \\  \addlinespace[0.4em]
Reverse KL & $\rKL(P\Vert \whP)$ & $ -\log u$ & $-1 - \log -t$ &  $-\whp(\vx)/p(\vx)$ & $1/\lambda$ \\ \addlinespace[0.4em]
$\chi^2$-Pearson & $\Dchi(P\Vert \whP) $ &  $(u-1)^2$ & $\frac{1}{4}t^2 + t$ & $2\left(p(\vx)/\whp(\vx)-1 \right)$ & $2/\lambda^3$ \\ \addlinespace[0.4em]
Total Variation    & $\TV$ & $\vert u-1\vert/2 $ & $t$ & $\sign\left(p(\vx)/\whp(\vx)-1 \right)/2 $ &  NA \\ \addlinespace[0.4em]
\bottomrule
\end{tabular*}
\end{sc}
\end{small}
\end{center}
\end{table*}

\subsection{\fdivs}
\label{subsec:fdiv}

Following the work by \citet{nowozin_f-gan_2016}, we consider the family of \fdivs to measure the difference between the true  distribution $P$  and the estimated  distribution $\whP$.  
Given a convex lower semi-continuous function $f:\reals^+ \rightarrow \reals$ satisfying $f(1)=0$, the \fdiv between two probability distributions $P$ and $\whP$ is defined as follows.
\begin{align}
\label{eq:fdiv}
	\Df (P\Vert\whP ) = \int_{\Xset} \whp(\vx) f\left(\frac{p(\vx)}{\whp(\vx)}\right) \d \vx.
\end{align}
Many well known  statistical divergences such as the Kullback Leibler divergence ($\KL$), the reverse Kullback Leibler ($\rKL$) or the Total Variation ($\TV$) are \fdivs (see Table~\ref{tab:fdiv}). $D_f$ also admits a dual variational form  \cite{nguyen_surrogate_2009}.
\begin{equation}
\begin{aligned}
\label{eq:dual}
	\Df(P \Vert \wh P)
     =\sup_{\T \in \cal T} \left( \E_{P}\left[T(\vx) \right] - \E_{\wh P}\left[ f^*(T(\vx))\right] \right),
\end{aligned}
\end{equation}
where \cal T is the set of measurable functions $\cal X \to \mathbb R$ and $f^* : \reals \to \reals$ is the convex conjugate (or Fenchel transform) of $f$ given by $f^*(t) = \sup_{u \in \reals} \left\{tu-f(u)\right\}$. 
We use $\Topt\in \cal T$ to denote the function that achieves the supremum in \eqref{eq:dual}.

Defining,
\begin{align}\label{eq:dual approx T}
    \Ddual = \E_{\vx\sim P}\left[T(\vx) \right] - \E_{\vx\sim \wh P}\left[ f^*(T(\vx))\right],
\end{align}
it is possible to train a generative model for $P$ by solving $\min_{F}\max_{\T}\Ddu_{f,T}(P \Vert \wh P)$, where $F$ and $T$ are represented by a neural network~\citep{nowozin_f-gan_2016}.

\subsection{Precision-Recall curve for generative models}
\label{subsec:PR}
\begin{figure}[ht]
	\subfigure[t][Distributions]{
		\includegraphics[width=0.46\linewidth]{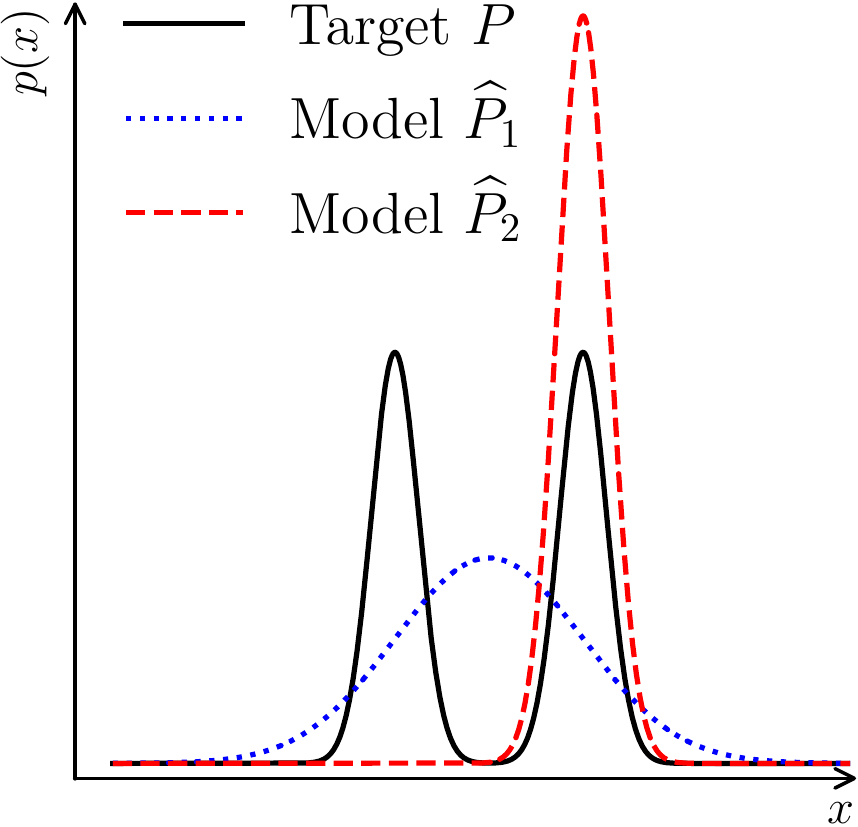}
  \label{fig:1DPRa}
	}
	\subfigure[t][$\partial\PRd$ curves]{
		\includegraphics[width=0.46\linewidth]{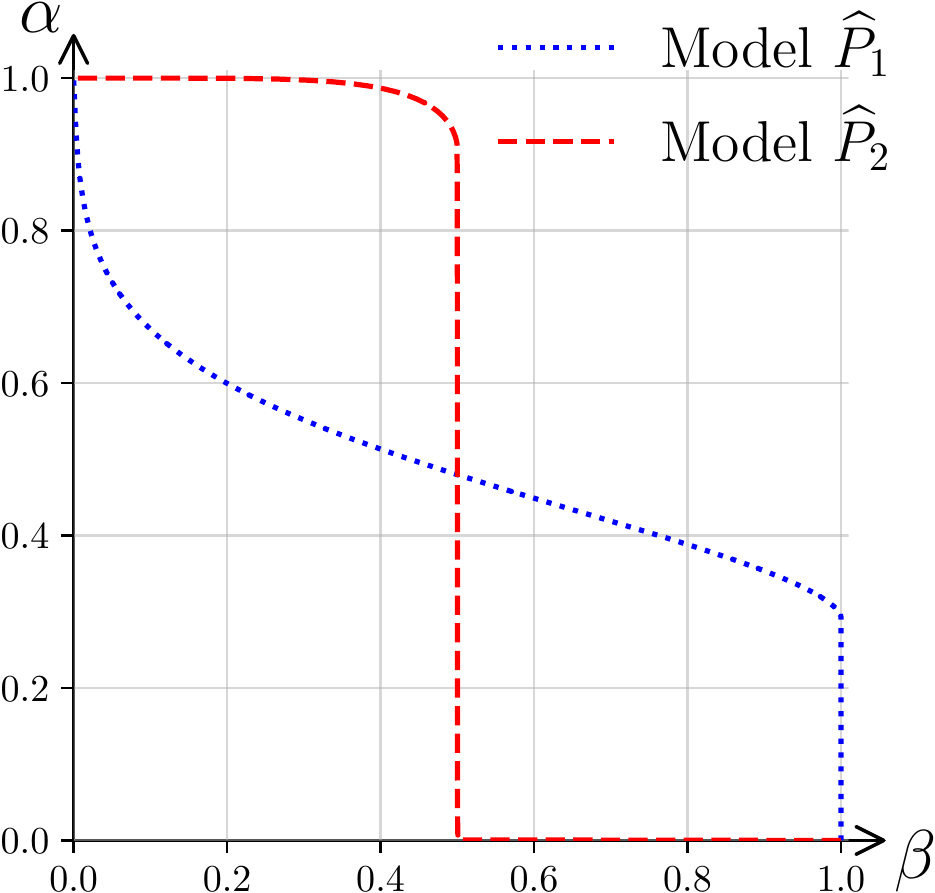}
  \label{fig:1DPRb}
	}
	\caption{PR curves for two models $\wh P_1$ and $\wh P_2$ of $P$. Figure~\ref{fig:1DPRa} shows $\wh P_1, \wh P_2$ and $P$. Figure~\ref{fig:1DPRa} shows PR curves for $\wh P_1, \wh P_2$ against $P$. $\whP_1$ has good recall since it covers both modes of $P$ but low precision since it generates points between the modes. $\whP_2$ has good precision since it does not generate samples outside of $P$ but low recall since it can generate samples from  only one mode.}
    \label{fig:1DPR}
\end{figure}

Typically, generative models are evaluated by a one-dimensional metric like the Inception Score \citep{salimans_improved_2016} or the Fr{\'e}chet Inception Distance \cite{heusel_gans_2017}, which are unable to distinguish between the distinct failure modes of low precision (i.e. failure to produce quality samples) and low recall (i.e. failure to cover all modes of $P$). The following definition introduced  by \citet{sajjadi_assessing_2018} and later extended by \citet{simon_revisiting_2019} remedies this problem.
\begin{definition}[PRD set, adapted from \citet{simon_revisiting_2019}]
For $P, \wh P\in \setP(\setX)$, {\em the Precision-Recall set} $\PRd$ is defined as the set of Precision-Recall pairs $(\alpha, \beta) \in \mathbb R^+~\times~\mathbb R^+$ such that there exists $\mu\in \setP(\setX)$ for which $P \ge \beta\mu$ and  $\wh P \ge \alpha\mu$.
The {\em precision-recall curve} (or PR curve) is defined as $\partial \PRd = \{ (\alpha, \beta) \in \PRd \mid \nexists (\alpha', \beta')\mbox{ with }\alpha' \ge \alpha\mbox{ and }\beta' \ge \beta \}$.
\end{definition}
\citet{sajjadi_assessing_2018} show that the PR curve is parametrised by $\lambda\in \mathbb R^+ \cup \{+\infty\}$  as follows:
\begin{equation}
\begin{split}
      &\partial\PRd = \left\{\alpha_\lambda(P\Vert\whP),\beta_\lambda(P\Vert\whP)~\vert~\lambda \in \mathbb R^+ \cup \{+\infty\} \right\}   \\
      &\quad\mbox{with }  \begin{cases}
          \alpha_\lambda(P\Vert\whP) = \int_{\Xset}\min\left(\lambda p(\vx), \whp(\vx)\right) \dx, \\
          \beta_\lambda(P\Vert\whP) = \int_{\Xset}\min\left( p(\vx), \whp(\vx)/\lambda\right) \dx. 
      \end{cases} 
\end{split}
\end{equation}
Note that $\beta_\lambda(P\Vert \wh P)  = \alpha_\lambda(P\Vert \wh P)/\lambda$.
We call $\lambda$ the {\em trade-off parameter} since it can be used to adjust the sensitivity to precision (or recall). 
An illustration of the PR curve is given in Figure~\ref{fig:1DPR} for a target distribution $P$ that is a mixture of two Gaussians, and two candidate models $\wh P_1$ and $\wh P_2$. 
We can see on Figure~\ref{fig:1DPR} that $\wh  P_1$ offers better results for large values of $\lambda$ (with high sensitivity to precision) whereas $\wh P_2$ offers better results for low values of  $\lambda$ (with high sensitivity to recall).

\section{Precision and Recall trade-off as a \fdiv}
In this section, we  formalise the link between precision-recall trade-off and \fdivs, and address Question~\ref{question: D lambda PR}. We will exploit this link in Section~\ref{sec: optimizing for PR} to train models that optimise a particular trade-off between precision and recall. 

\subsection{Precision-Recall as an \fdiv}\label{subsec: DPR defs and link}

We start by introducing the {\em PR-Divergence} as follows. 

\begin{definition}[PR-divergence]
Given a trade-off parameter $\lambda \in \mathbb R^+\cup +\infty$, the {\em PR-divergence} (denoted by $\DPR$) is defined as the $f_\lambda$-divergence for $f_\lambda : \mathbb R^+ \to \mathbb R$ defined as $$f_\lambda :  u \mapsto \max(\lambda u, 1) - \max(\lambda, 1),$$
for $\lambda\in \mathbb R^+$ and $f_\lambda :  u \mapsto 0$ for $\lambda = +\infty$.
\end{definition}
Note that $f_\lambda$ is continuous, convex, and satisfies  $f_\lambda(1)=0$ for all $\lambda$. 
A graphical representation  of $f_\lambda$ can be found in Appendix~~\ref{app:prop:DPR}.
The following proposition gives some properties of $\DPR$.

\begin{proposition}[Properties of the PR-Divergence]\label{prop:DPR}
	\begin{itemize}
		\item The Fenchel conjugate $f_\lambda^*$ of $f_\lambda$ is defined on $\dom\left(f_\lambda\s\right) = \left[ 0, \lambda\right]$ and given by,
			\begin{align}
				f_\lambda\s\left(t\right) = \begin{cases}t/\lambda & \mbox{ if }\,\lambda\leq1, \\ t/\lambda+\lambda-1 & \mbox{otherwise.}\end{cases}
			\end{align}
		\item The discriminator $\Topt$ that achieves the supremum in the variational representation of $\DPR(P \Vert \wh P)$ is
			\begin{align}
				\Topt(\vx) = \lambda \sign\left\{\frac{p(\vx)}{\whp(\vx)} - 1\right\}.
			\end{align}
            \item $\DPR(\whP \Vert P) = \lambda 
    \fullDPR{\frac{1}{\lambda}}(P\Vert \whP)$.
            \item $ \fullDPR{1}(P\Vert \whP) = \TV(P\Vert \whP)/2$.
		\end{itemize}
\end{proposition}

Having defined the PR-divergence, we can now show that precision and recall w.r.t $\lambda$ can be expressed as a function of the divergence between $P$ and $\wh P$.

 \begin{theorem}[Precision and recall as a function of $\DPR$]\label{thm:DPR}
    Given  $P,\wh P\in \setP(\setX)$ and $\lambda  \in \mathbb R^+ \cup \{+\infty\}$, the PR curve $\partial\PRd$ is related to the PR-divergence $\DPR(P\Vert \whP)$ as follows.
\begin{align}
     \alpha_\lambda(P\Vert \wh P) &= \min(1, \lambda)-\DPR(P\Vert \whP).
\end{align}
\end{theorem}
A direct consequence of Theorem~\ref{thm:DPR} is that we can now train the parameters of a model $F_\theta$ to maximise precision and recall by minimising the corresponding PR-divergence.
\begin{align}
    \argmax_\theta ~ \alpha_\lambda(P\Vert \wh P_{F_\theta}) = \argmin_\theta ~\DPR(P\Vert \whP_{F_\theta}). 
\end{align}

In other terms, training a model to minimise $\DPR$ means that the model will specifically focus the set trade-off. One might wonder what is the trade-off, ie the function $a_\lambda$ minimised when the model is trained on the $\KL$. 
\subsection{Relation between PR-divergences and other \fdivs}
In the previous subsection, we showed that for each trade-off parameter $\lambda$, there exists an \fdiv that corresponds to optimising for it. This raises the converse question of what trade-off is achieved by optimizing for an arbitrary \fdiv. We answer this by showing in the following theorem that  any \fdiv can be expressed as a weighted sum of  PR-divergences. 

\begin{theorem}[\fdiv as weighted sums of PR-divergences]\label{thm: weighted PRD}
For any  $P,\wh P\in \setP(\setX)$ supported on all of $\setX$ and any $\lambda  \in \mathbb R^+ $,
\begin{equation}\label{thm:fdivPR}
    \Df(P\Vert\whP) = \int^{M}_{m}\frac{1}{\lambda^3}f''\left(\frac{1}{\lambda}\right)\DPR(P\Vert\whP) \d\lambda,
    \end{equation}
where $m =\min_{\setX}\left(\frac{\whp(\vx)}{p(\vx)}\right)$ and $M =\max_{\setX}\left(\frac{\whp(\vx)}{p(\vx)}\right)$.
\end{theorem}


Combining Theorem~\ref{thm: weighted PRD} with Theorem~\ref{thm:DPR} we have the following relation that captures the implicit precision-recall trade-off made by minimising an arbitrary \fdiv.
\begin{align}
    \argmin_\theta \Df(P\Vert\whP)
    = \argmax_\theta \int_{m}^{M}\frac{1}{\lambda^3}f''\left(\frac{1}{\lambda}\right)\alpha_\lambda(P\Vert\whP) \d\lambda,
\end{align}
In the following corollary, we apply Theorem~\ref{thm: weighted PRD} to compare $\KL$ and $\rKL$.
\begin{corollary}[$\KL$ and $\rKL$ as an average of $\DPR$]\label{cor:KLrKLPR}
The $\KL$ Divergence and the $\rKL$ can be written as weighted average of PR-Divergence $\DPR$:
  \begin{align}
      \KL(P\Vert\whP)=\int_{m}^{M}\frac{1}{\lambda^2} \DPR(P\Vert\whP)\d\lambda,
   \end{align}
    \begin{align}
       \rKL(P\Vert\whP)=\int_{m}^{M} \frac{1}{\lambda} \DPR(P\Vert\whP)\d\lambda.
    \end{align}
\end{corollary}

As we can see in this Corollary, both $\KL$ and $\rKL$ can be decomposed into a sum of PR-divergences terms $\DPR$, each weighted with  $1/\lambda^2$ and $1\lambda$ respectively.
Hence $\KL$ gives more weight to small $\lambda$ (more sensitive to precision) than $\rKL$.
This explains the {\em mode covering} behaviour observed in Normalizing Flows trained with $\KL$. Comparatively, the $\rKL$ assigns more weight to terms with a large lambda and less weight to the terms with a small lambda, leading to the  {\em mode covering} behaviour empirically observed with flows trained with the $\rKL$ \cite{midgley_bootstrap_2022}.

\section{Optimising for specific Precision-Recall trade-offs}\label{sec: optimizing for PR}

In this section, we address Question~\ref{question: how to train model} by proposing a model that achieves a specified precision-recall trade-off. In light of Theorem~\ref{thm:DPR}, a natural approach would be to use the framework of \citet{nowozin_f-gan_2016} and train an $f_\lambda$-GAN to minimize the dual variational form of $\DPR$ for any given $\lambda$, as follows.
\begin{align*}
    &\min_{F}\max_{\T}\Ddu_{f_\lambda,T}(P \Vert \wh P)\\
    &=\min_{F}\max_{\T} \E_{\vx\sim P}\left[T(\vx) \right] - \E_{\vx\sim \wh P}\left[ f^*_\lambda(T(\vx))\right],
\end{align*}
where both $F$ and $T$ are parametrized by neural networks. 
As noted in \citet{nowozin_f-gan_2016}, we need to use an output activation function $\sigma_{\lambda}$ on $T(x)$ so that $\sigma(T(x))\in dom(f_\lambda^*)$. From Proposition~\ref{prop:DPR}, we have that the domain of $f^*_\lambda$ is bounded (equal to $[0,\lambda]$) unlike most other popular \fdivs including $\KL$ and $rKL$, $\chi^2$, Jensen-Shannon and Hellinger, except for the case of Total Variation. Due to this, we are forced to choose an output activation function like the sigmoid that suffers from vanishing gradients. Because of this, we find that $f_\lambda$-GAN is notoriously hard to train and performs poorly, not unlike the case of training $f$-GAN with the Total Variation metric.

We overcome this problem by using the primal form of \fdiv shown in \eqref{eq:fdiv} to minimise $\DPR$  rather than the dual variational form in \eqref{eq:dual}, as explained in the following subsection. 



\subsection{Minimising the primal estimation of $\DPR$}
\label{subsec:primal_estimate}
\begin{figure}\label{fig: model arch}
    \centering
    \includegraphics[width=\linewidth]{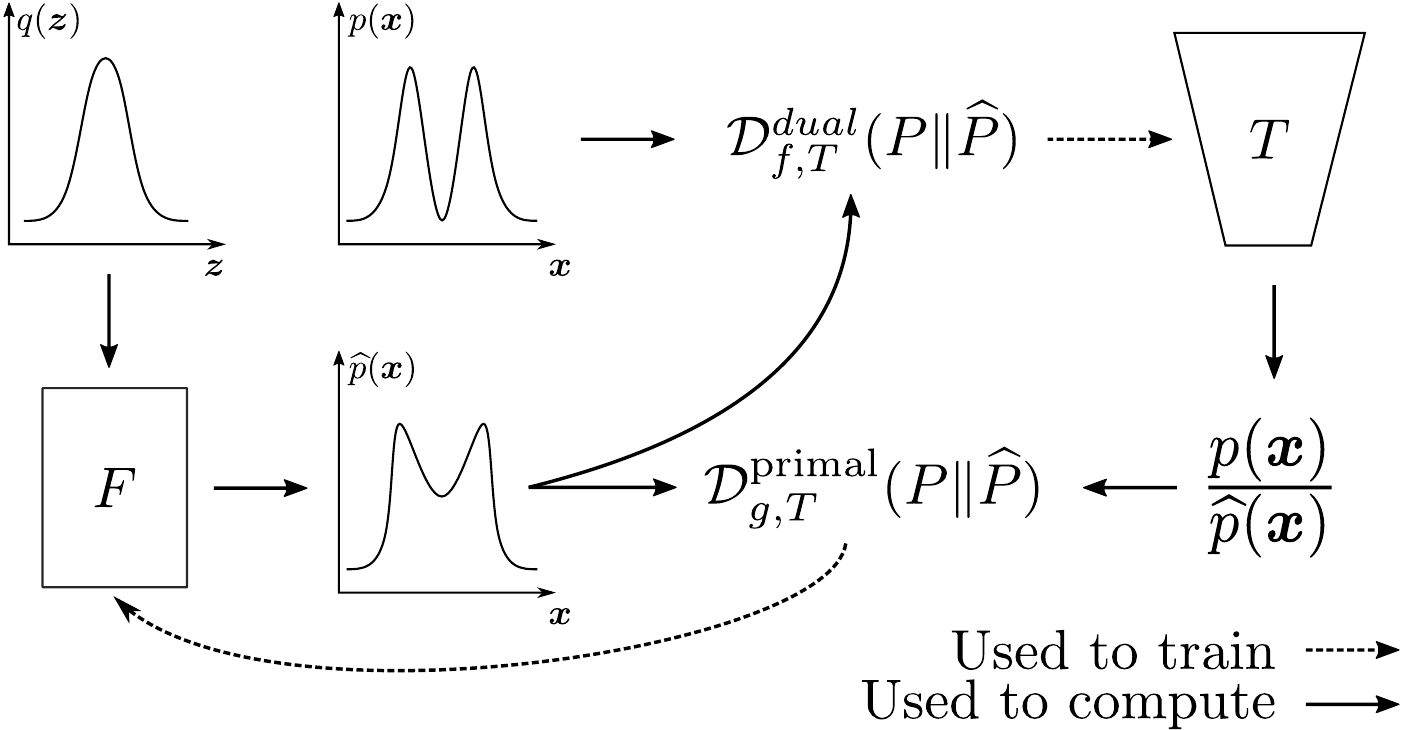}
    \caption{Model architecture for training a Normalising Flow $F$ that minimises $\D_g$ using a discriminator $T$ that minimises $\D_f$.}
    \label{fig:training process}
\end{figure}

As explained before, using the the $f$-GAN approach to minimise the dual variational form of $\DPR$ is not a feasible approach. 
In this subsection, we introduce a novel approach to instead train a Normalizing Flow to minimise $\DPR$. 
We do this by using the primal form of $f$-divergence, given by $\Df (P\Vert\whP ) = \int_{\Xset} \whp(\vx) f\left(\frac{p(\vx)}{\whp(\vx)}\right) \d \vx$. Computing $D_f$ with the primal form requires the likelihood ratio $p(x)/\wh p(x)$. For this, we use the $\Topt$ from the dual form of $D_f$. From the dual form, we have the following.
\begin{align*}
    \Df(P \Vert \wh P)
     =\sup_{\T \in \cal T} \int_{\setX} \left( p(x)T(x) - \wh P(x)f^*(T(x))\right) dx.
\end{align*}
Differentiating the expression inside the integral with respect to $T(x)$, we get the following.
\begin{align}\label{eq: primal likelihood}
    \frac{p(x)}{\wh p(x)} = \nabla f^*(\Topt(x)).
\end{align}
We use the preceding equation to propose the following definition for a {\em primal estimate} of \fdiv.


\begin{definition}[Primal estimate  $\D^{\mathrm{primal}}_{g, T}$]
Let $P, \wh P\in \setP(\setX)$.
For functions $T:\mathcal{X}\to \mathbb{R}$ and $f, g: \mathbb{R}^+\to \mathbb{R}$, we define the primal estimate $\D^{\mathrm{primal}}_{g, T}$ of the divergence $\cal D_g$ as follows.
\begin{align}
    \D^{\mathrm{primal}}_{g, T}(P \Vert \wh P)=\int_{\Xset} \whp(\vx)g\left(r(\vx)\right)\d\vx,
\end{align}
where  $r:\mathcal{X}\rightarrow\mathbb{R}^+$ is given by, $r(\vx) = \nabla f\s (T(\vx))$.  
 \end{definition}

Crucially, when $\Topt$ is as given in \eqref{eq: primal likelihood}, we have $\D^{\mathrm{primal}}_{g, T} = \D_g$. 
Also note that $r(x)$ acts as the estimate for the likelihood $p(x)/\wh p(x)$.
With this, we propose the following scheme for training a Normalizing Flow $F$ that minimises $\D_g$ by using a discriminator $T$ that minimises $\D_f$.
\begin{itemize}
    \item The discriminator $T$ is trained to minimise $\D_f$ with the dual variational form.
    \begin{align*}
        &\argmin_T \Ddual \\
        &= \argmin_T \E_{P}\left[T(\vx) \right] - \E_{\wh P}\left[ f^*(T(\vx))\right].
    \end{align*}   
    \item The generator, a Normalizing Flow  $F$, is trained to maximise $\D_g$ with the primal form.
    \begin{align*}
        &\argmin_T \D^{\mathrm{primal}}_{g, T} \\
        &= \argmin_T \E_{\wh P}\left[ g(\nabla f^*(T(\vx)))\right].
    \end{align*}     
\end{itemize}
Note that for $T = \Topt$, we have $\Ddual = \D_f $, and consequently, $\D^{\mathrm{primal}}_{g, T} = \D_g$.
Figure~\ref{fig: model arch} shows the model architecture and Algorithm~\ref{alg:training process} depicts the training process.

Finally, in order to train a generator to minimise $\DPR$, we use $g = f_\lambda$ and an $f$ for which $dom(f^*)$ is not bounded, for example $KL$ or $rKL$ or Jensen-Shannon divergence. In the following subsection, we discuss how the choice of $f$ affects the model training.


\begin{algorithm}[tb]
   \caption{Training }
   \label{alg:training process}
\begin{algorithmic}
   \STATE {\bfseries Input:} Flow $F$, Discriminator  $T$, Dataset $\DD$
   \FOR{ epoch $e=1, \dots, E$}
   \STATE $\mathcal{L}_{d}, \mathcal{L}_{p} \leftarrow 0, 0$
   \FOR{$\vx_{\mathrm{real}}  \in \DD$}
   \STATE Generate $\vx_{\mathrm{fake}} = F\inv(\vz)$ with $\vz \sim \N(\vzero_d, \mI_d)$
   \STATE $ \mathcal{L}_{d} \leftarrow \mathcal{L}_{d} + T(\vx_{\mathrm{real}}) - f\s(T(x_{\mathrm{fake}}))$
   \STATE $\mathcal{L}_{d} \leftarrow \mathcal{L}_{d} + g\left(\nabla f\s(T(\vx_
   {\mathrm{fake}})\right)$
   \ENDFOR
   \STATE Update $\T$ to maximise $\mathcal{L}_{d}$
   \STATE Update $F$  to minimise $\mathcal{L}_{p}$
   
   \ENDFOR
\end{algorithmic}
\end{algorithm}

\subsection{Choosing the right $f$ to train T}
To use the proposed model for optimizing for PR-divergences, we fix $g = f_\lambda$. This still leaves us with the flexibility to choose $f$ for which $dom(f^*)$ is not bounded. The choice of $f$ affects the estimate $r(x) = \nabla f^*(T(x))$ of the likelihood $p(x)/\wh p(x)$. In the following theorem, we show that the Bregman divergence associated with $f$ place a crucial role in the goodness of approximation of the dual form $\Ddual$.



\begin{theorem}[Error of the estimation of an \fdiv under the dual form.]
For any discriminator $T:\mathcal{X}\rightarrow\mathbb{R}$ and $r\left(\vx\right)=\nabla f\s (T(\vx))$,
\begin{align}
D_{f}(P\Vert \whP)-D_{f,T}^{\mathrm{dual}}(P\Vert \whP)=\mathbb{E}_{\whP}\left[\breg_{f}\left(r(\vx),\frac{p(\vx)}{\whp(\vx)}\right)\right].
\end{align}
\label{thm:breg}
\end{theorem}

In the next theorem, we bound the approximation error of $\mathcal{D}_{g,T}^{\mathrm{primal}}(P\Vert \whP)$.

\begin{theorem}[Bound on the estimation of an \fdiv using another \fdiv]\label{thm: f g divs}
Let $f, g: \mathbb{R}^+\to \mathbb{R}$ be such that $f$ is $\mu$-strongly
convex and $g$ is $\sigma$-Lipschitz. For discriminator $T:\mathcal{X}\rightarrow\mathbb{R}$, let $r\left(\vx\right)=\nabla f\s (T(\vx))$. If 
\begin{align*}
    \Df(P\Vert \whP)-\Ddual\le\epsilon,
\end{align*}
then,
\begin{align*}
    \left\vert \mathcal{D}_{g}(P\Vert \whP)-\mathcal{D}_{g,T}^{\mathrm{primal}}(P\Vert \whP)\right\vert \le\sigma\sqrt{\frac{2\epsilon}{\mu}}.
\end{align*}
\label{thm:boundedestimation}
\end{theorem}

\begin{figure}[H]
    \centering
    \includegraphics[width=\linewidth]{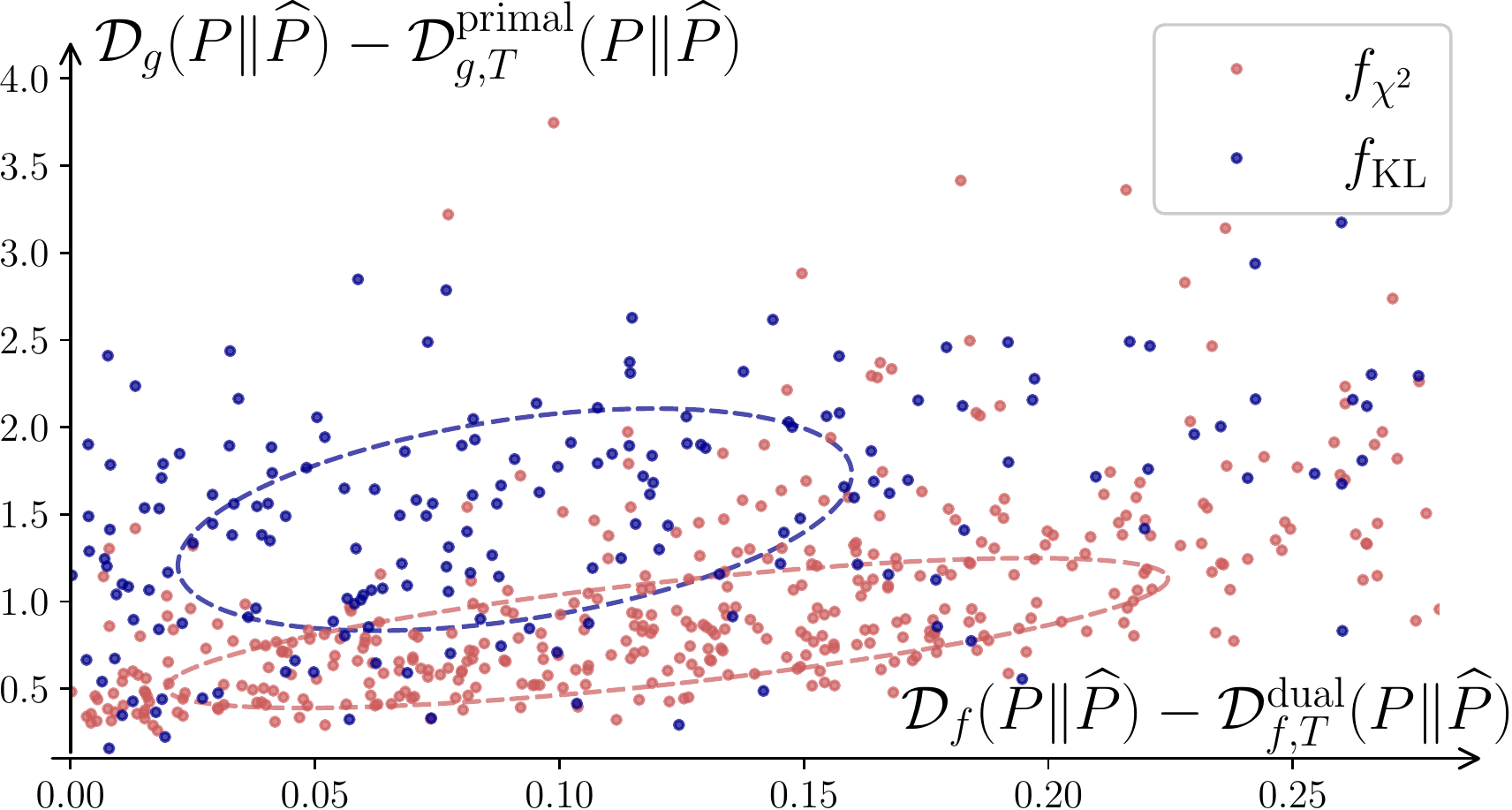}
    \caption{200 different $P$ and $\whP$ composed of a mixture of 15 Gaussians have been used to train two discriminators: one on $\KL$ and one on $\Dchi$. For each discriminator, the distance between $\D_g$ and its primal and, the distance between $\D_f$ and its dual have been reported. The lower the point is on the $y$-axis,  the better the estimation is. The red dots are discriminators train with $f_{\chi^2}$ and in blue with $f_{\mathrm{KL}}$. The ellipsoid represent the Mahalanobis of each set. }
    \label{fig:ch2KL}
\end{figure}

Note that $g = f_\lambda$ is indeed Lipschitz for any $\lambda$, and $\D_f$ can be chosen so that $f$ is strongly convex (for example, the $\chi^2$ divergence). Hence, Theorems~\ref{thm:breg} and \ref{thm: f g divs} together provide theoretical support for the convergence of our proposed model. In Figure~\ref{fig:ch2KL}, we show empirical evidence for the convergence of primal and dual forms.

\begin{figure}[t]
\label{fig:prfig}
\centering
    \subfigure[c][ $\KL$]{\includegraphics[width=0.45\linewidth]{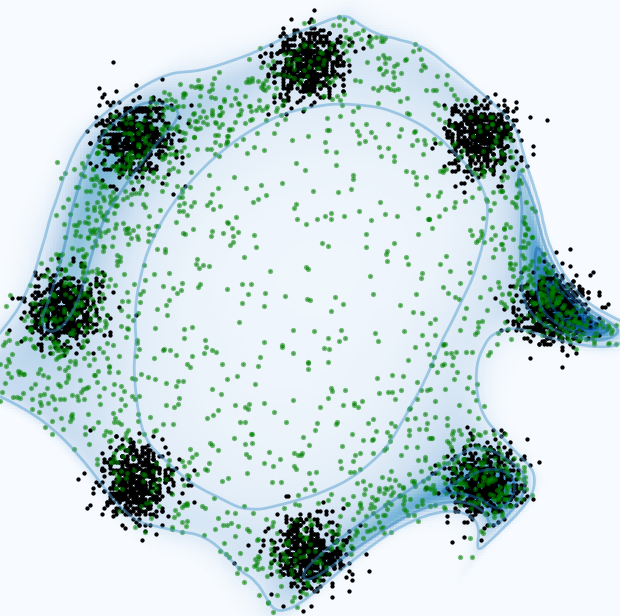} \label{fig:kl}}
    \subfigure[c][ $\rKL$]{\includegraphics[width=0.45\linewidth]{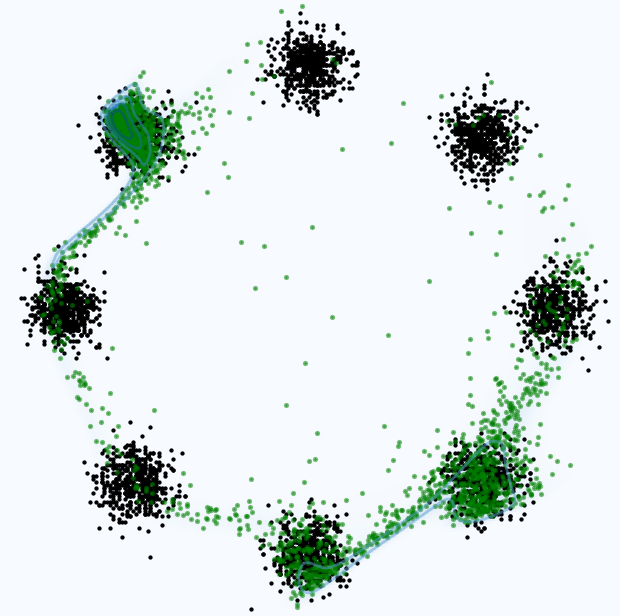}  \label{fig:rkl}}
    \subfigure[c][ AUC]{\includegraphics[width=0.45\linewidth]{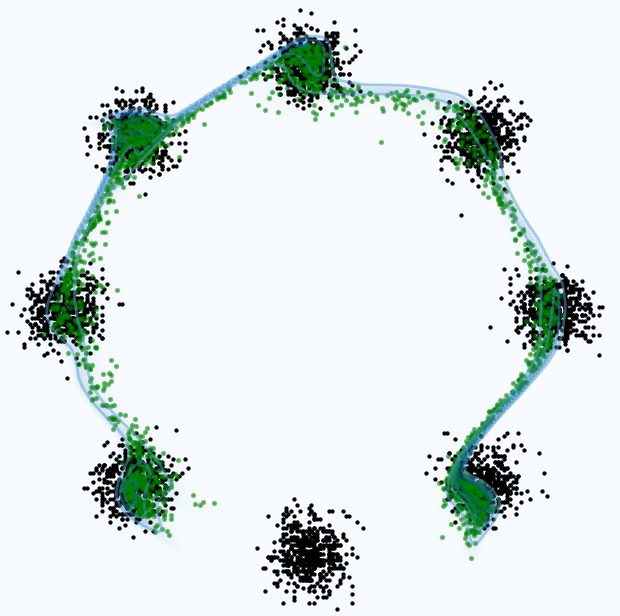}  \label{fig:auc}}
    \caption{RealNVP trained on a two dimensional 8 Gaussians Mixtures. Samples from $P$ are represented in black while samples drawn from $\whP$ are green. The log-likelihood of $\whP$ is represented as the blue levels, the darker the higher. The models have been trained  to minimise (from left to right) the $\KL$, the $\rKL$ and the Aera Under the Curve (AUC). The corresponding PR-curves are plotted in Figure~\ref{fig:PRGaus}.}
\end{figure}

\section{Practical considerations}

\subsection{A new approach to compute $\partial \PRd$}
To evaluate our model, the frontier $\partial \PRd$ must be computed. The original work of \citet{sajjadi_assessing_2018} proposes a clustering method to estimate every point $\left(\alpha(\lambda), \beta(\lambda)\right)$ but was proven to fail estimate packed data points \cite{kynkaanniemi_improved_2019}. A more recent approach \cite{simon_revisiting_2019} proposed to train a binary classifier to make the difference between points from $P$ and $\whP$ and then use false positive and false negative ratios to estimate the frontier. Since we have proven that the frontier can be estimated by the primal of $\DPR$ and this for every $\lambda \in \reals^+$, we can simply use the estimated ratio $\frac{p(\vx)}{p(\vx)}$ obtained with the Discriminator $\T$. Therefore, with $r(\vx) = \nabla f\s(\T(\vx))$ as the estimated ratio, the estimated precision on the frontier can be written as. By using $\Dchi$ to train $\T$, we certify a bounded error on the estimated precision in opposition to the crossentropy (i.e $\KL$) used in the previous approach. 

\subsection{Training for optimal AUC}

Inspired by the work from classification, we can compute and even train the model to optimise the {\em Area Under the Curve} (AUC). This will result in a model that achieve good  performance across the entire range of  possible trade-off values $\lambda$ instead of maximising the performance for one particular trade-off.  The AUC can be computed with its expression given in Proposition~\ref{prop:AUC} and the model can be trained on this loss. 

\begin{proposition}[AUC under the $\partial \PRd$]
The aera under the curve is: 
\begin{align}
    \mathrm{AUC} = \int_{0}^{+\infty}\alpha_\lambda(P\Vert \whP)^2\d\lambda
\end{align}
\label{prop:AUC}
\end{proposition}
\todo{Add conclusion sentence to this subsec}

\section{Experiments}
In this section, we report on a series of experiments conducted to illustrate the benefits of using the PR-divergence. A first series of experiments is conducted on a 2D dataset (8 gaussians) and the other uses higher dimensional image datasets, MNIST and CelebA. We train RealNVPs \cite{rezende_variational_2016} for the first dataset and GLOW~\cite{kingma_glow_2018} for the high dimensional datasets. For every model, the discriminator $T$ is trained to maximise $\Dchi^{\mathrm{dual}}(P\Vert\whP)$, then the models are trained to minimise different losses estimated with $T$ as described in Section~\ref{subsec:primal_estimate}. 

\paragraph{8 Gaussian dataset}
Figures~\ref{fig:l01},~\ref{fig:l1} and~\ref{fig:l10} present models trained on the 8 Gaussians dataset, with $\DPR$ using $\lambda = 0.1$, $\lambda = 1$ and $\lambda = 10$.
As we can see, increasing $\lambda$ dramatically improves precision at the cost of recall and vice-versa.   
Figure~\ref{fig:kl},~\ref{fig:rkl} and~\ref{fig:auc}, presents  models that have been trained to minimise respectively $\KL$, $\rKL$ and the AUC. The mass covering and the mode seeking behaviours of $\KL$ and $\rKL$ are clearly demonstrated on this figures. Notice however, that thanks to the flexibility of the PR-divergence, adjust $\lambda$ and choose every model in between these extreme behaviours. 

The corresponding PR-curves are plotted in Figure~\ref{fig:prfig}. The two green curves are the frontier  $\partial \PRd$ for the $\KL$ and the $\rKL$ and can be set as the reference. For $\lambda = 1$, the model is in between both curves, worse recall than $\KL$ and worse precision than  $\rKL$ but has a better precision than $\KL$ and a better recall than  $\rKL$. With the high value of $\lambda$, we have set a relatively high importance on the precision, and the model performs better than $\rKL$ in terms of precision. Finally, the AUC model is not always as good as more specialised models, but has the best AUC and the best $\TV$.
\begin{figure}[H]
\centering
\subfigure[$\lambda=\frac{1}{5}$]{\includegraphics[width = 0.3\linewidth]{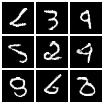}}
\subfigure[$\lambda=1$]{\includegraphics[width = 0.3\linewidth]{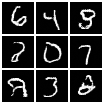}}
\subfigure[$\lambda=5$]{\includegraphics[width = 0.3\linewidth]{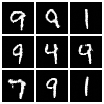}}
\subfigure[PR-curves]{\includegraphics[width = 0.65 \linewidth]{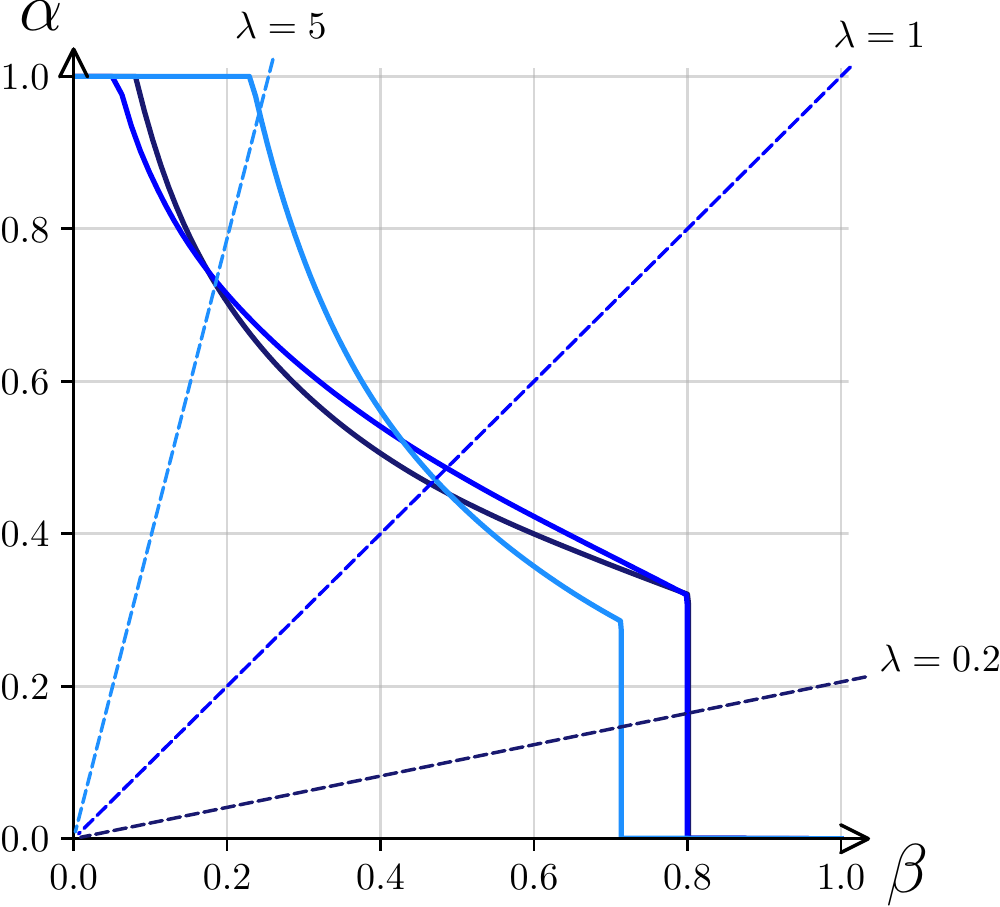}}
\caption{Glow models trained on MNIST. Small batches of samples are given for the different models.}
\label{fig:mnist}
\end{figure}
\paragraph{MNIST and CelebA} To test the framework on high dimensional image dataset, we have trained different multi-scale Glow models  on the dataset MNIST \cite{yann_lecun_mnist_2010} and a CelebA $64\small{\times}64$ \cite{liu_deep_2015}. Three models for each data set have trained minimising $\DPR$ with  $\lambda = 1/5$, $\lambda = 1$ and $\lambda = 5$. Samples of generated images are showed in Figure~\ref{fig:mnist} for MNIST and Figure~\ref{fig:celeba}. (Larger batches are available in Appendix~\ref{app:xp}.) For MNIST, we can see that for values of $\lambda\leq 1$, the generate images have a poor quality and it is represented on the PR-curve as both curves quickly decrease. However, the samples for the model trained with $\lambda = 10$ generates samples with low variance (mostly from class $1$, $9$ $8$, $6$ and $4$) but with a better quality. The PR-curve is thus greater for high values of $\lambda$ and lower for low values of $\lambda$. On CelebA, most normalizing flows have trouble generating quality samples with various background. Here we can see that by setting the trade-off on recall, the model generates a wide spectrum of background but with poor quality faces. On the contrary,  by setting the trade-off on precision, the model standardise the  and improve the precision on the faces.

\begin{figure}[H]
\centering
\subfigure[$\lambda=\frac{1}{5}$]{\includegraphics[width = 0.3\linewidth]{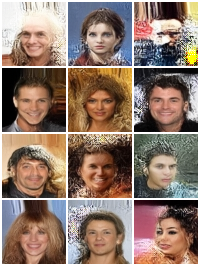}}
\subfigure[$\lambda=1$]{\includegraphics[width = 0.3\linewidth]{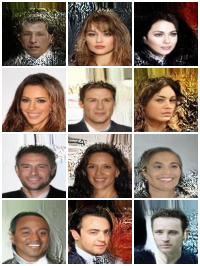}}
\subfigure[$\lambda=5$]{\includegraphics[width = 0.3\linewidth]{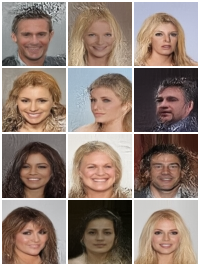}}
\caption{Glow models trained on MNIST. Small batches of samples are given for the different models.}
\label{fig:celeba}
\end{figure}

\section{Conclusion}

In this paper, we present a method for training Normalizing Flows using a new Precision/Recall (PR) divergence within the framework of f-divergences. Our approach offers a unique advantage over other divergences as it allows for explicit control of the precision-recall trade-off in generative models. The PR-divergence results in models that range from extreme mode seeking (high precision) to extreme weight covering (high recall), as well as more balanced models that may be more suitable for various applications. Our framework also provides new insights into other well-known divergences, such as the KL and reverse KL. Our experiments indicate that models trained for high precision (using the reverse KL) often have high precision but poor recall. Given that detecting low recall in a model is more difficult than detecting low precision through visual inspection of model samples, we suspect that existing evaluation methodologies have overemphasised high precision models. As such, we hope that this research will ultimately lead to the development of better tools for evaluating the overall quality of generative models.

\section*{Acknowledgements }
We are grateful for the grant of access to computing resources  at the IDRIS Jean Zay cluster under allocation No. AD011011296 made by GENCI.

\bibliography{references}

\bibliographystyle{icml2023}

\newpage
\appendix
\renewcommand\thefigure{\thesection.\arabic{figure}}
\onecolumn
\section{Appendix A. }

\subsection{Proof for Theorem~\ref{thm:DPR}}
\label{app:thm:DPR}
	We have to prove that $\alpha(\lambda)$ can be written as a function of an \fdiv for any $\lambda\in\reals^+$.
	First we can develop the expression of $\alpha(\lambda)$:
\begin{align} 
	\alpha(\lambda) &= \int_{\setX} \min \left(\lambda p(\vx), \whp(\vx) \right)\dx \\
			&= \int_{\setX} \whp(\vx) \min\left(\lambda \frac{p(\vx)}{\whp(\vx)}, 1 \right)\dx
\end{align}
For this integral to be considered as an \fdiv, we need $f$ to be first convex lower semi-continuous and then to satisfy $f(1)=0$. However, for every $a,b\in\reals$, the $\min$ satisfies $\min(a,b) = a+b - \max(a,b)$. Therefore, 
\begin{align} 
	\alpha(\lambda) &= \int_{\setX} \whp(\vx) \left[\lambda \frac{p(\vx)}{\whp(\vx)} + 1 - \max\left(\lambda \frac{p(\vx)}{\whp(\vx)}, 1 \right)\right]\dx \\
			&=  \lambda \int_{\setX}p(\vx)\dx + 1 - \int_{\setX}\max\left(\lambda \frac{p(\vx)}{\whp(\vx)}, 1 \right)\dx\\
			&= \lambda +1 - \int_{\setX}\whp(\vx)\max\left(\lambda \frac{p(\vx)}{\whp(\vx)}, 1 \right)\dx
\end{align}
Thus, we can take $f(u) = \max(\lambda u , 1) - \max(\lambda, 1)$ such that $f(1)=0$. The precision becomes: 
\begin{align} 
	\alpha(\lambda) &= \lambda +1 - \int_{\setX}\whp(\vx)f(\left(\frac{p(\vx)}{\whp(\vx)} \right) - \max(\lambda, 1)\int_{\setX}\whp(\vx)\dx \\
			&= \min(\lambda, 1)- \int_{\setX}\whp(\vx)f(\left(\frac{p(\vx)}{\whp(\vx)} \right)\dx = \min(\lambda, 1) - \DPR(P, \whP).
\end{align}
Consequently, $\alpha(\lambda)$ can be written as a function of an \fdiv $\DPR$ with $f(u) = \max\left(\lambda u, 1\right) - \max\left(\lambda, 1\right)$.

\subsection{Proof of Proposition~\ref{prop:DPR}}
\label{app:prop:DPR}
\begin{figure}[H]
	\subfigure[Function $f(u)$ for different values of $\lambda$.]{\includegraphics[width = 0.5\linewidth]{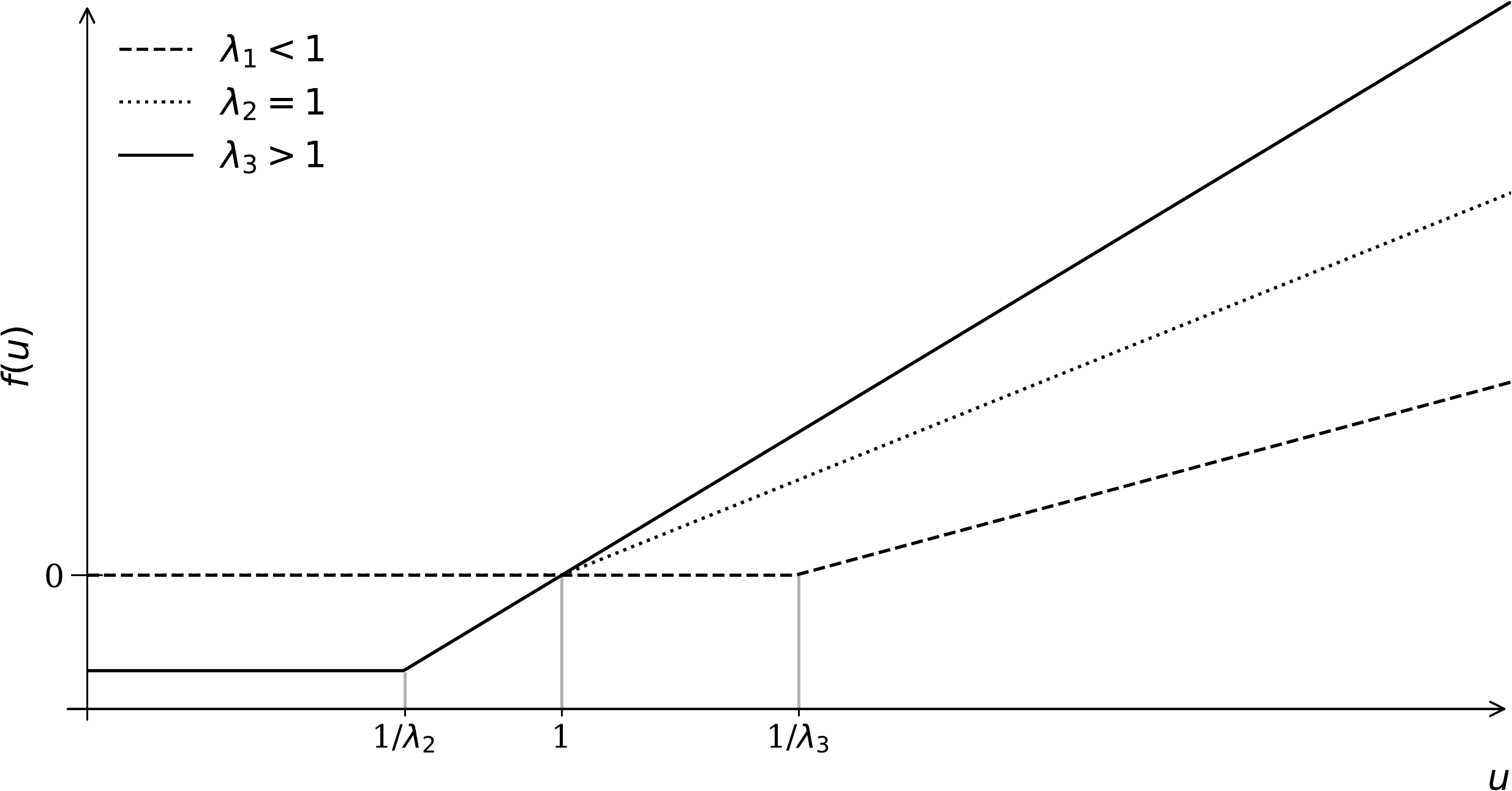} \label{fig:fpr}}
	\subfigure[Function $u\mapsto ut - \max( \lambda u, 1)$ for values of $t$ between $0$ and $\lambda$.]{\includegraphics[width=0.5\linewidth]{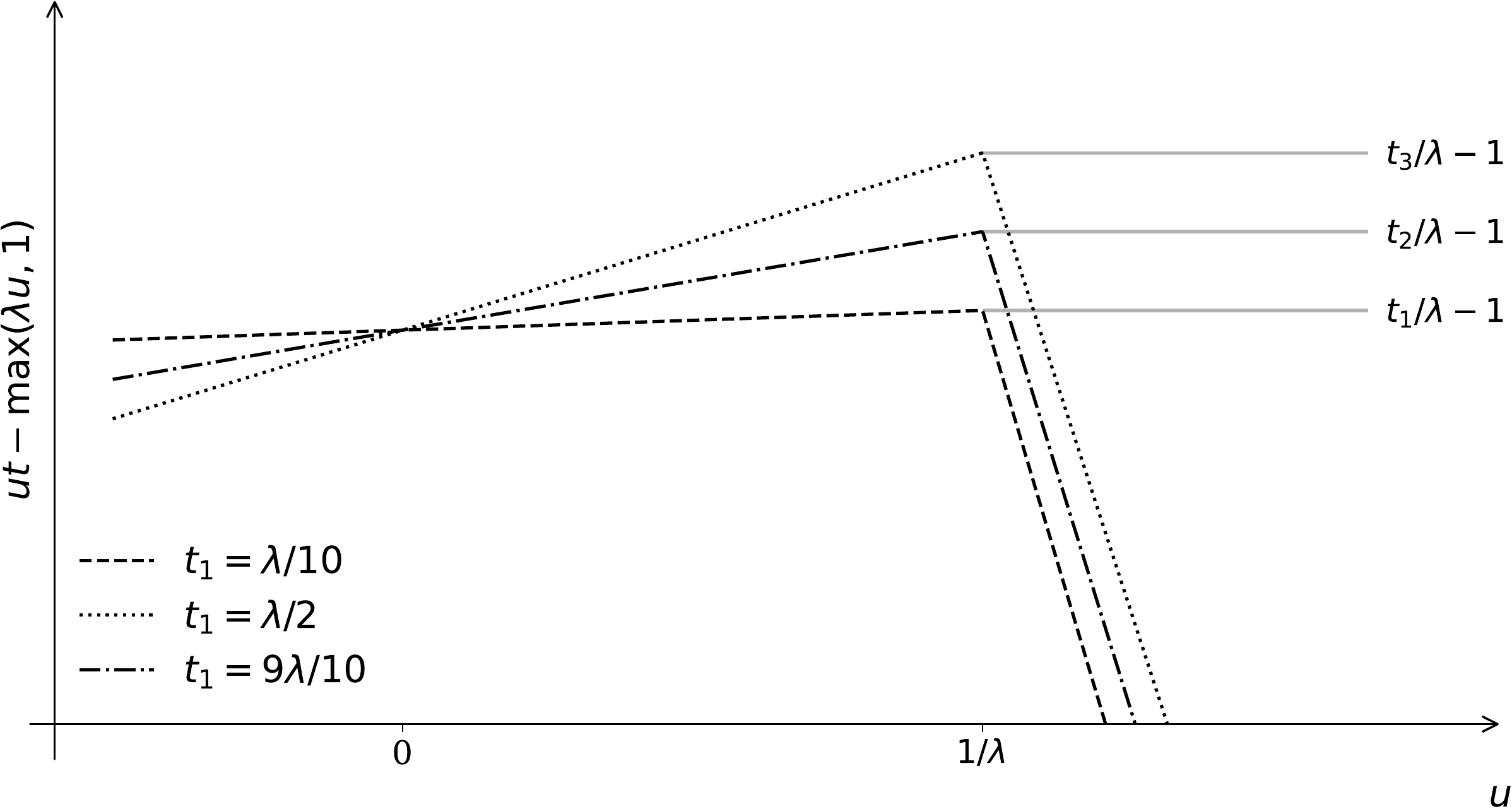}\label{fig:fenchelpr}}
\end{figure}

 If the generator function $f$ of the Precision-Recall Divergence is 
$f(u) = \max(\lambda u , 1) - \max(\lambda, 1)$ then its Fenchel conjugate function is:  
\begin{align} 
		f\s(t)  = \sup_{u\in \dom(f)}\left\{tu - f(u)\right\} = \max(\lambda, 1)  + \sup_{u\in\reals^+}\left\{tu - \max\left(\lambda u, 1\right)\right\}
\end{align}
If $t>\lambda$ or $\lambda<0$, then the $\sup_{u\in\reals^+}\left\{tu - \max\left(\lambda u, 1\right)\right\} =\infty$ for respectively $u\rightarrow \infty$ and  $u\rightarrow -\infty$. The domain of $f\s$ is thus restricted to $\left[0, \lambda\right]$. 
Thus for $0\leq t\leq \lambda$, the supremum is obtained for $u = 1/\lambda$ since $0$ is in the sub-differential of the function in $1/\lambda$ as Figure~\ref{fig:fenchelpr}. 
Consequently the Fenchel conjugate of $f$ is:
\begin{align} 
	\forall t\in\left[0,  \lambda\right], \quad		f\s(t)  = \max(\lambda, 1)  + t\lambda - 1 = \begin{cases} t/\lambda & \mbox{if } \lambda \leq 1, \\ t/\lambda -1+\lambda & \mbox{otherwise. } \end{cases}
\end{align}
Finally, the optimal discriminator $\Topt$ by taking the derivative of $f$ in $\frac{p(\vx)}{\whp(\vx)}$, we get:
\begin{align}
	\Topt(\vx) = \nabla f \left(\frac{p(\vx)}{\whp(\vx)}\right) = \begin{cases} \lambda & \si \frac{p(\vx)}{\whp(\vx)}\leq 1/\lambda , \\ 0 & \mbox{otherwise}. \end{cases}
\end{align}

Then we can compute the compute the reverse $\DPR$:
 \begin{align}
     \DPR(\whP\Vert P)& = \int_{\setX}p(\vx) f_{\lambda} \left(\frac{\whp(\vx)}{p(\vx)}\right)\d \vx \\
     & = \int_{\setX}\max(\lambda \whp(\vx), p(\vx)) - p(\vx)\max(\lambda, 1)\, \d \vx \\
     & = \lambda\left(  \int_{\setX}\max( \whp(\vx), p(\vx)/\lambda )\partial \vx - \max(1, 1/\lambda )\right)  \\
    & = \lambda\int_{\setX}\whp(\vx)\max( 1, \frac{p(\vx)}{\whp(\vx)}/\lambda ) -\whp(\vx) \max(1, 1/\lambda )\, \d \vx \\
    & = \lambda  \int_{\setX}\whp(\vx) f_{1/\lambda} \left(\frac{p(\vx)}{\whp(\vx)}\right)\d \vx \\
    & = \lambda  \fullDPR{\frac{1}{\lambda}}(P\Vert \whP).
 \end{align}
 With this results, we can show that : 
 \begin{align}
     \TV(P\Vert \whP) &= \int_{\setX} \left\vert p(\vx) - \whp(\vx)\right\vert \d\vx \\
      &=\int_{\setX} \max( p(\vx) - \whp(\vx), 0) + \max( \whp(\vx) - p(\vx), 0)  \d\vx
 \end{align}
 Then since $\fullDPR{1}(P\Vert \whP) = \int_{\setX}\max( \whp(\vx), p(\vx)) - p(\vx)\d \vx = \int_{\setX}p(\vx)\max( \whp(\vx)\d \vx $ and  $\fullDPR{1}(P\Vert \whP) = \fullDPR{1}(\whP\Vert P)$,  we have:
  \begin{align}
     \TV(P\Vert \whP) &= \fullDPR{1}(P\Vert \whP) + \fullDPR{1}(\whP\Vert \P) \\
      &=2\fullDPR{1}(P\Vert \whP).
 \end{align}
\subsection{Proof of Theorem~\ref{thm:breg}}
\label{app:thm:breg}

From now on, assume the support of $P$ and $\whP$ coincide. For any $T:\mathcal{X}\rightarrow\mathbb{R}$, 
\begin{align*}
\Ddual& =\mathbb{E}_{x\sim P}\left[d\left(x\right)\right]-\mathbb{E}_{x\sim Q}\left[f\s\left(d\left(x\right)\right)\right]\\
 & =\mathbb{E}_{x\sim Q}\left[\frac{p(\vx)}{\whp(\vx)}d\left(x\right)-f\s\left(d\left(x\right)\right)\right]
\end{align*}

Let $\Topt\in\arg\sup \D_{f,T}^{\mathrm{dual}}(P\Vert\whP)$. For any $T:\mathcal{X}\rightarrow\mathbb{R}$

\begin{align*}
\Df(P\Vert\whP)-D_{f,T}^{\mathrm{dual}}(P\Vert\whP) & =\D_{f,\Topt}^{\mathrm{dual}}(P\Vert\whP)-\D_{f,T}^{\mathrm{dual}}(P\Vert\whP)\\
 & =\mathbb{E}_{\whP}\left[\frac{p(\vx)}{\whp(\vx)}\left(\Topt(\vx)-T(\vx)\right)-f\s\left(\Topt(\vx)\right)+f\s\left(T(\vx)\right)\right]
\end{align*}

It is known that for all $x\in\mathcal{X}$ we have $\nabla f\s(\Topt(\vx))=\frac{p(\vx)}{\whp(\vx)}$:

\begin{align*}
\Df(P\Vert\whP)-\D_{f,T}^{\mathrm{dual}}(P\Vert\whP) & =\mathbb{E}_{\whP}\left[\nabla f\s(\Topt(\vx))\left(\Topt(\vx)-T(\vx)\right)-f\s\left(\Topt(\vx)\right)+f\s\left(T(\vx)\right)\right]
\end{align*}

Recall that for any continuously differentiable strictly convex function$f$, the Bregman divergence of $f$ is $\breg_{f}\left(a,b\right)=f(a)-f(b)-\left\langle \nabla f(b),a-b\right\rangle $.
So we have

\begin{align*}
\Df(P\Vert\whP)-\D_{f,T}^{\mathrm{dual}}(P\Vert\whP) & =\mathbb{E}_{\whP}\left[\breg_{f\s}\left(T(\vx),\Topt(\vx)\right)\right]
\end{align*}

Let us now use the following property: $\breg_{f}\left(a,b\right)=\breg_{f\s}\left(a\s,b\s\right)$
where $a\s=\nabla f(a)$ and $b\s=\nabla f(b)$.

\begin{align*}
\Df(P\Vert\whP)-\D_{f,T}^{\mathrm{dual}}(P\Vert\whP) & =\mathbb{E}_{\whP}\left[\breg_{f}\left(\nabla f\s(T(\vx)),\nabla f\s(\Topt(\vx))\right)\right]\\
 & =\mathbb{E}_{\whP}\left[\breg_{f}\left(\nabla f\s(T(\vx)),\frac{p(\vx)}{\whp(\vx)}\right)\right]
\end{align*}

Let us define $r\left(\vx\right)=\nabla f\s T(\vx)$
as our estimator of $p(\vx)/\whp(\vx)$. So finally, we have

\begin{align*}
\Df(P\Vert\whP)-\D_{f,T}^{\mathrm{dual}}(P\Vert\whP) & =\mathbb{E}_{\whP}\left[\breg_{f}\left(r(\vx),\frac{p(\vx)}{\whp(\vx)}\right)\right]
\end{align*}

\subsection{Proof of Theorem~\ref{thm:boundedestimation}}
\label{app:thm:boundedestimation}
Now assume that $f$ is $\mu$-strongly convex, then $\breg_{f}(a,b)\ge\frac{\mu}{2}\left\Vert a-b\right\Vert ^{2}$
If $\mathbb{E}_{\whP}\left[\breg_{f}\left(r(\vx),\frac{p(\vx)}{\whp(\vx)}\right)\right]\le\epsilon$
and if $f$ is $\mu$-strongly convex, then

\begin{align}\label{eq: thm bdd est}
\mathbb{E}_{\whP}\left[\left(r(\vx)-\frac{p(\vx)}{\whp(\vx)}\right)^{2}\right]\le\frac{2\epsilon}{\mu}.
\end{align}

Consider an arbitrary f-divergence $\D_{g}(P\Vert\whP)=\int g\left(\frac{dP}{\d\whP}\right)\d\whP$. Define $\D_{g,T}^{\mathrm{primal}}(P\Vert \whP)=\int g\left(r(\vx)\right)\d\whP$. Then, 
\begin{align*}
\left|\D_{g}(P\Vert\whP)-\D_{g,T}^{\mathrm{primal}}(P\Vert \whP)\right| & =\left|\mathbb{E}_{\whP}\left[g\left(\frac{p(\vx)}{\whp(\vx)}\right)-g\left(r(\vx)\right)\right]\right|\\
 & \le\mathbb{E}_{\whP}\left[\left|g\left(\frac{p(\vx)}{\whp(\vx)}\right)-g\left(r(\vx)\right)\right|\right]\\
 &\stackrel{(a)}{\le} \mathbb{E}_{\whP}\left[\sigma\left|e(x)\right|\right]\\
 &= \sigma\mathbb{E}_{\whP}\left[\left|e(x)\right|\right]\\
  &\stackrel{(b)}{\le} \sigma\sqrt{\mathbb{E}_{\whP}\left[e(x)^2\right]}\\
   &\stackrel{(c)}{\le} \sigma \sqrt{\frac{2\epsilon}{\mu}},
\end{align*}
where $(a)$ follows from the $\sigma$-Lipschitz assumption on $g$, $(b)$ follows from Jensen's inequality and finally, $(c)$ follows from equation~\eqref{eq: thm bdd est}.


\subsection{Proof of Theorem~\ref{thm:fdivPR}}
\label{app:thm:fdivPR}
Let $c:\reals^+ \mapsto \reals$ be a $\mathcal{C}^2$ function and take $u_{\min}$ and $u_{\max}$. The goal is to express $f(u)$ for all $u\in[u_{\min}, u_{\max}]$ as a weighted average of $f^{\mathrm{PR}}_\lambda$:
\begin{align}
	\forall u\in\reals_*^+, \int_{1/u_{\max}}^{1/u_{\min}}c''(\lambda)f^{\mathrm{PR}}_\lambda(u
)\d\lambda =  \int_{1/u_{\max}}^{1/u_{\min}}c''(\lambda)\left[\max(\lambda u, 1)-\max\left(\lambda, 1\right)\right]\d\lambda
\end{align}
First, let us assume that $u_{\min}\leq1$ and $u_{\max}\geq1$, then the terms can be decomposed and the integral split to evaluate the $\max$:
\begin{align}
	\int_{1/u_{\max}}^{1/u_{\min}}c''(\lambda)f^{\mathrm{PR}}_\lambda(u
)\d\lambda &= \int_{1/u_{\max}}^{1/u_{\min}}c''(\lambda)\max(\lambda u, 1)\d\lambda -\int_{1/u_{\max}}^{1/u_{\min}}c''(\lambda)\max\left(\lambda, 1\right)\d\lambda \\
		&= \begin{multlined}[t]
		    \int_{1/u_{\max}}^{1/u}c''(\lambda)\max(\lambda u, 1)\d\lambda +\int_{1/u}^{1/u_{\min}}c''(\lambda)\max(\lambda u, 1) \d\lambda \\
-\int_{1/u_{\max}}^{1}c''(\lambda)\max\left(\lambda, 1\right)\d\lambda  -\int_{1}^{1/u_{\min}}c''(\lambda)\max\left(\lambda, 1\right)\d\lambda 
		\end{multlined}\\
            &=  \int_{1/u_{\max}}^{1/u}c''(\lambda)\d\lambda +\int_{1/u}^{1/u_{\min}}c''(\lambda)\lambda u\d\lambda
-\int_{1/u_{\max}}^{1}c''(\lambda)\d\lambda  -\int_{1}^{1/u_{\min}}c''(\lambda)\lambda\d\lambda .
\end{align}
By integrating by parts, we have: $\int_{1/u_{\max}}^{1/u_{\min}}c''(\lambda)\lambda\d\lambda = \left[c'(\lambda)\lambda\right]_{1/u_{\max}}^{1/u_{\min}} - \int_{1/u_{\max}}^{1/u_{\min}}c'(\lambda)\d\lambda$ so it satisfies:
\begin{align}
	\int_{1/u_{\max}}^{1/u_{\min}}c''(\lambda)f^{\mathrm{PR}}_\lambda(u
)\d\lambda &=  \begin{multlined}[t] \int_{1/u_{\max}}^{1/u}c''(\lambda)\d\lambda + u \left[c'(\lambda)\lambda \right]_{1/u}^{1/u_{\min}} - u \int_{1/u}^{1/u_{\min}}c'(\lambda)\d\lambda\\
 -\int_{1/u_{\max}}^{1}c''(\lambda)\d\lambda - \left[c'(\lambda)\lambda \right]_{1}^{1/u_{\min}} +  \int_1^{1/u_{\min}}c'(\lambda)\d\lambda\end{multlined}\\
            &=  \begin{multlined}[t]\left[c'(\lambda)\right]_{1/u_{\max}}^{1/u} + u \left[c'(\lambda)\lambda \right]_{1/u}^{1/u_{\min}} - u \left[c(\lambda)\right]_{1/u}^{1/u_{\min}}  \\ -\left[c'(\lambda)\right]_{1/u_{\max}}^{1} - \left[c'(\lambda)\lambda \right]_{1}^{1/u_{\min}} +  \left[c(\lambda)\right]_1^{1/u_{\min}}\end{multlined}\\
            &= \begin{multlined}[t] c'\left(\frac{1}{u}\right) - c'(0) + u c'\left(\frac{1}{u_{\min}}\right) - u c'\left(\frac{1}{u}\right)\frac{1}{u} - u c\left(\frac{1}{u_{\min}}\right) +uc\left(\frac{1}{u}\right)-c'(1) \\
            + c'(0)  -  c'\left(\frac{1}{u_{\min}}\right)\frac{1}{u_{\min}} + c'(1)\times1 +c\left(\frac{1}{u_{\min}}\right) - c(1) \end{multlined} \\
            &= \left[c'\left(\frac{1}{u_{\min}}\right)\frac{1}{u_{\min}} - c\left(\frac{1}{u_{\min}}\right)\right]\left(u-1\right) + uc\left(\frac{1}{u}\right)-c(1).
\end{align}
We would like $\int_{1/u_{\max}}^{1/u_{\min}}c''(\lambda)f^{\mathrm{PR}}_\lambda(u
)\d\lambda$ to be equal to $f$ between $u_{\min}$ and $u_{\max}$. But since two \fdivs generated by $f$ and $g$ are equals if there is a $c\in\reals$ such that $f(u)=g(u)+c(u-1)$, the Divergence generated by $\int_{1/u_{\max}}^{1/u_{\min}}c''(\lambda)f^{\mathrm{PR}}_\lambda(u
)\d\lambda$ is equal to the divergence generated by $uc\left(\frac{1}{u}\right)-c(1)$. Therefore, we require $c$ to satisfy: 
\begin{align*}
    \forall u\in[u_{\min}, u_{\max}], \quad f(u) = uc\left(\frac{1}{u}\right)-c(1).
\end{align*}
By differentiating with respect to $u$, we have:
\begin{align}
    f'(u) =  \lim_{\lambda \rightarrow \infty}\left[c'(\lambda)\lambda - c(\lambda)\right] + c\left(\frac{1}{u}\right) -\frac{1}{u}c'\left(\frac{1}{u}\right).
\end{align}
And finally:
\begin{align}
    f''(u) &=  -\frac{1}{u^2}c\left(\frac{1}{u}\right) +\frac{1}{u^2}c'\left(\frac{1}{u}\right) +\frac{1}{u^3}c''\left(\frac{1}{u}\right) \\
    & = \frac{1}{u^3}c''\left(\frac{1}{u}\right).
\end{align}
Consequently, with $\lambda = 1/u$, we have that:
\begin{align}
    c''(\lambda) = \frac{1}{\lambda^3}f''\left(\frac{1}{\lambda}\right).
    \label{eq:coeffPR}
\end{align}
With such a results, with $m=\min_{\setX}(\frac{\whp(\vx)}{p(\vx)})$ and $M=\max_{\setX}(\frac{\whp(\vx)}{p(\vx)})$, we can write any \fdiv as:
\begin{align*}
    \Df(P\Vert \whP) & = \int_{\setX} \whp(\vx) f\left(\frac{p(\vx)}{\whp(\vx)}\right)\dx  \\
    &=  \int_{\setX} \whp(\vx) \int_{m}^{M} \frac{1}{\lambda^3}f''\left(\frac{1}{\lambda}\right) f^{\mathrm{PR}}_\lambda\left(\frac{p(\vx)}{\whp(\vx)}\right) \d\lambda \dx \\
     &=  \int_{m}^{M}\int_{\setX}  \frac{1}{\lambda^3}f''\left(\frac{1}{\lambda}\right) \whp(\vx)  f^{\mathrm{PR}}_\lambda\left(\frac{p(\vx)}{\whp(\vx)}\right) \d\lambda \dx \\
      &= \int_{m}^{M} \frac{1}{\lambda^3}f''\left(\frac{1}{\lambda}\right)\left[\int_{\setX}  \whp(\vx)  f^{\mathrm{PR}}_\lambda\left(\frac{p(\vx)}{\whp(\vx)}\right)\dx  \right] \d\lambda \\
        &=  \int_{m}^{M} \frac{1}{\lambda^3}f''\left(\frac{1}{\lambda}\right)\DPR(P\Vert\whP) \d\lambda \\
\end{align*}

\subsection{Proof of Corollary~\ref{cor:KLrKLPR}}
\label{app:cor:KLrKLPR}
In particular for the $\KL$, $f(u)=u\log u$, therefore $f''(u) = 1/u$ which gives: 
\begin{align}
    \KL(P\Vert\whP)=\int_{m}^{M}\frac{1}{\lambda^2} \DPR(P\Vert\whP)\d\lambda.
\end{align}
And for the $\rKL$ we can either use Equation~\ref{eq:coeffPR} with $f(u)-\log u$ or use the fact that $\DPR(P\Vert\whP) = \lambda\DRP(\whP\Vert P)$:
\begin{align}
    \rKL(P\Vert\whP)=\int_{m}^{M}\frac{1}{\lambda} \DPR(P\Vert\whP)\d\lambda.
\end{align}

\subsection{Proof of Proposition~\ref{prop:AUC}}
\label{app:prop:AUC}
The AUC can be computed by integrating with respect to an angle $\theta$ on the quadrant:

\begin{align}
    \mathrm{AUC} = \int_0^{\pi/2} r^2(\theta)\d\theta.
\end{align}
Therefore with $\lambda = \tan \theta$, we have $r(\theta) = \alpha_{\tan(\theta)}(P\Vert\whP) / \cos\theta$. Thus:
\begin{align}
    \mathrm{AUC} &= \int_0^{\pi/2}\frac{ \alpha_{\tan(\theta)}(P\Vert\whP)^2 }{\cos^2\theta}\d\theta\\
    & =   \int_0^{+\infty} \alpha_{\lambda}(P\Vert\whP)^2 \cos^2\theta\d\lambda \quad \mbox{with } \quad \frac{\d\theta}{\cos^2\theta} = \partial \lambda.  
\end{align}

\begin{figure}[H]
    \centering
    \includegraphics[width = 0.3 \linewidth]{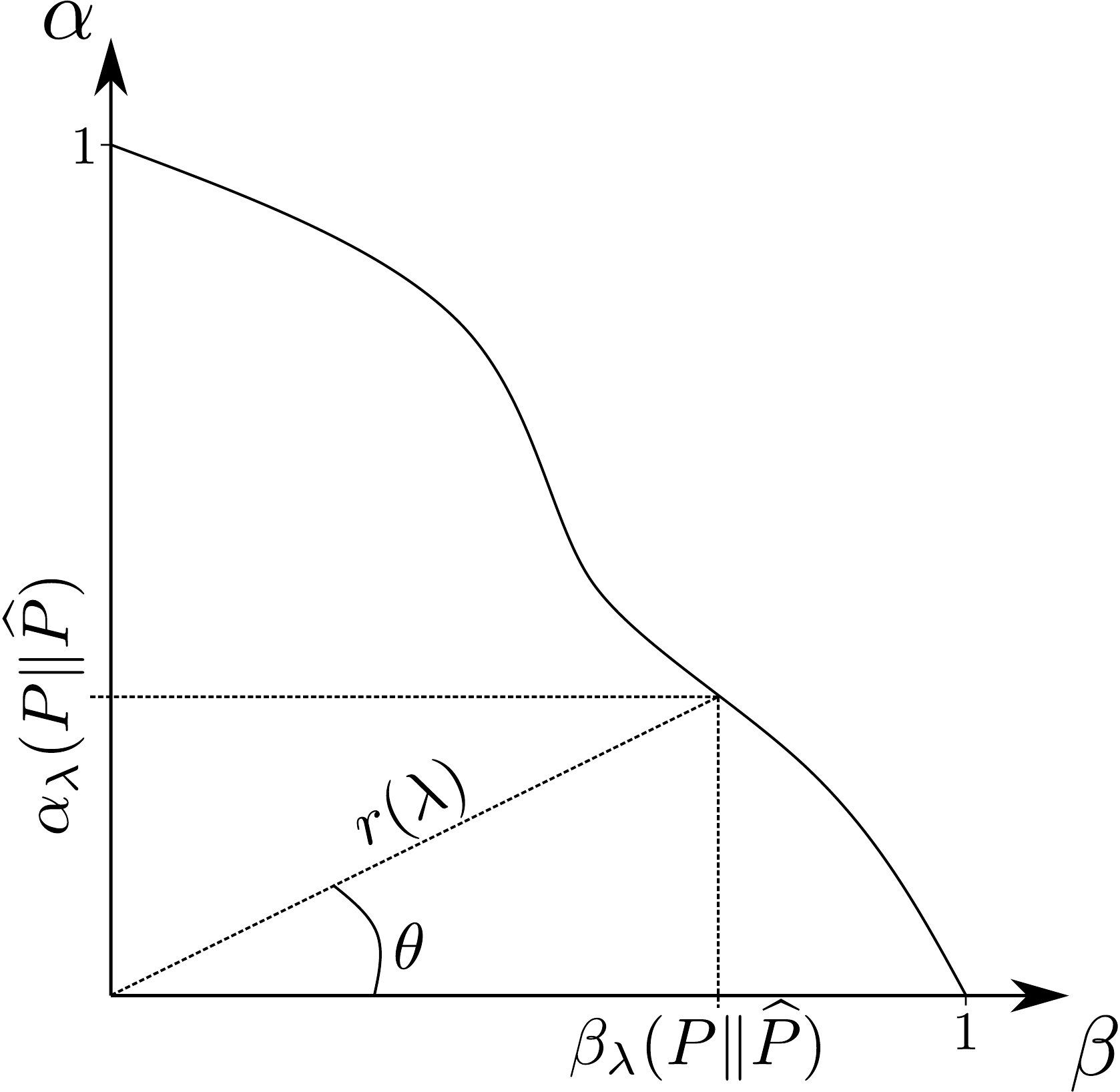}
    \caption{Illustration of the change of variable to compute the AUC. Instead of parametrising the frontier $\partial \PRd$ with $\lambda\in\reals \cup \left\{\infty\right\}$, we take $\theta\in\left[0, \frac{\pi}{2}\right]$ with $\lambda = \tan \theta$.}
    \label{fig:my_label}
\end{figure}
\section{Experiments}
\label{app:xp}
\begin{figure}[H]
\centering
\subfigure[$\lambda=\frac{1}{5}$]{\includegraphics[width = 0.3\linewidth]{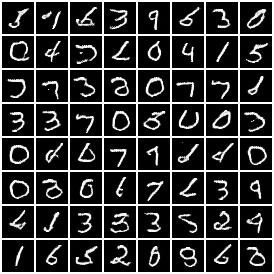}}
\subfigure[$\lambda=1$]{\includegraphics[width = 0.3\linewidth]{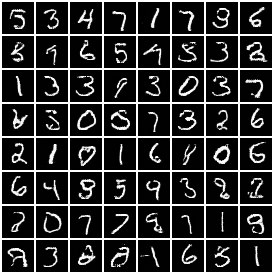}}
\subfigure[$\lambda=5$]{\includegraphics[width = 0.3\linewidth]{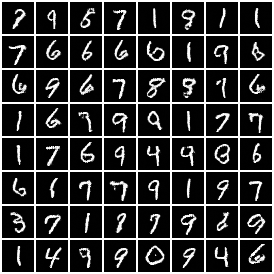}}
\end{figure}
\begin{figure}[H]
\centering
\subfigure[$\lambda=\frac{1}{5}$]{\includegraphics[width = 0.3\linewidth]{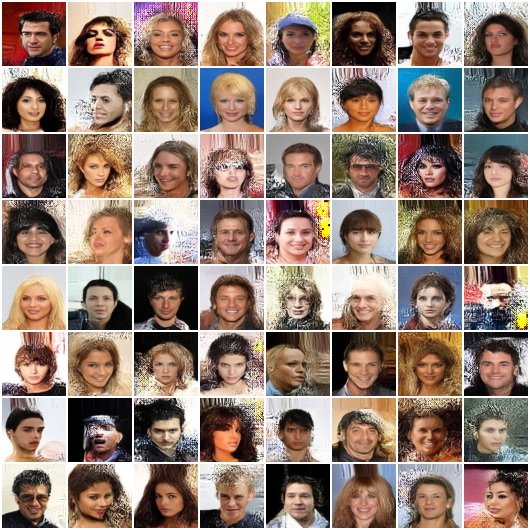}}
\subfigure[$\lambda=1$]{\includegraphics[width = 0.3\linewidth]{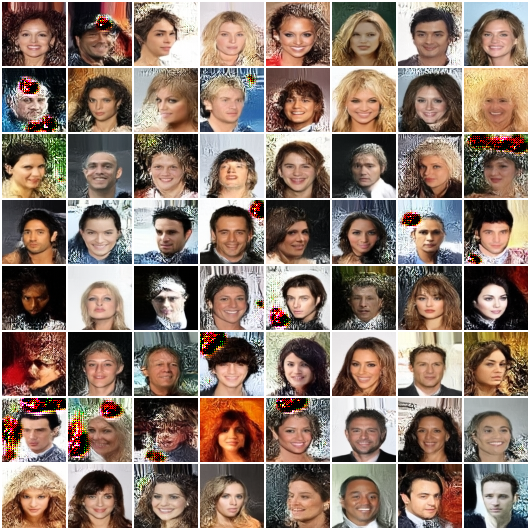}}
\subfigure[$\lambda=5$]{\includegraphics[width = 0.3\linewidth]{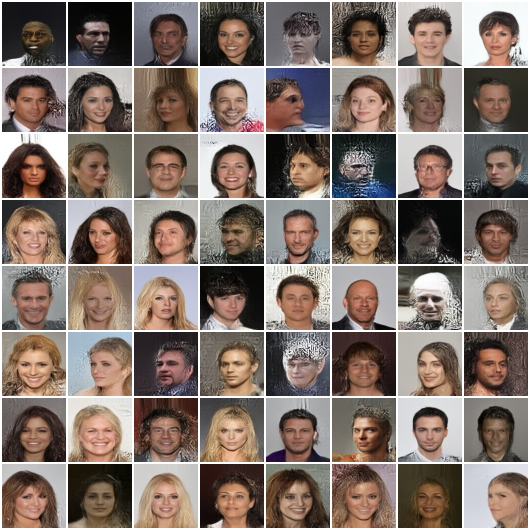}}
\caption{Large batch of samples for different Glow models trained on MNIST ((a), (b) and (c)) and CelebaA ((d), (e) and (f)). Models have been trained to minimise $\DPR$ with different $\lambda$ given in caption. The lower $\lambda$ is, the greater the model will value the recall, while precision is highly considered  for high values of $\lambda$.}  
\end{figure}
The details of the experiments are described in Table~\ref{app:xptab}

\begin{table}[H]
    \centering
    \begin{tabular}{|c|c|c|c|}
    \hline
        Dataset & Model & \#Parameters $F$ & \#Parameters $D$ \\
        \hline
        8 Gaussians & RealNVP & $540$k & $659$k  \\
        MNIST & Glow & $85$M & $1.7$M \\
        CelabA  & Glow  & $188$M & $13.2$M \\\hline
    \end{tabular}
    \caption{Details of the models for the different experiments. 
}
    \label{tab:xptab}
\end{table}


\end{document}